\documentclass[a4paper,fleqn]{cas-dc}

\usepackage[authoryear]{natbib}
\makeatletter
\g@addto@macro\normalsize{%
  \setlength{\abovedisplayskip}{2pt plus 1pt minus 1pt}%
  \setlength{\belowdisplayskip}{8pt plus 1pt minus 1pt}%
  \setlength{\abovedisplayshortskip}{2pt plus 1pt minus 1pt}%
  \setlength{\belowdisplayshortskip}{8pt plus 1pt minus 1pt}%
  \setlength{\jot}{1pt}%
}
\makeatother

\usepackage{silence}
\usepackage{float}

\makeatletter
\def\printorcid{}
\makeatother

\def\tsc#1{\csdef{#1}{\textsc{\lowercase{#1}}\xspace}}
\tsc{WGM}
\tsc{QE}

\begin{document}
\let\WriteBookmarks\relax
\def\floatpagepagefraction{1}
\def\textpagefraction{.001}

\shorttitle{} 

\shortauthors{}
 
\title[mode = title]{NavEYE: Vision-Centered Multi-Sensor Fusion-Based Situational Awareness System for Intelligent Surface Vehicles}  



%
 
\author[1,2,3]{Ryan Wen Liu}

\cormark[1]


\ead{wenliu@whut.edu.cn}


\credit{Methodology, Data curation, Validation, Software, Funding acquisition, Supervision, Writing - original draft, Writing - review \& editing}

\affiliation[1]{organization={Sanya Science and Education Innovation Park, Wuhan University of Technology},
                city={Sanya},
                postcode={572000}, 
                state={Hainan},
                country={China}}
\affiliation[2]{organization={School of Navigation, Wuhan University of Technology},
                city={Wuhan},
                postcode={430063}, 
                state={Hubei},
                country={China}}
\affiliation[3]{organization={State Key Laboratory of Maritime Technology and Safety},
                city={Wuhan},
                postcode={430063}, 
                state={Hubei},
                country={China}}

\author[1,2,3]{Junxiong Liang}




\credit{Methodology, Validation, Visualization, Writing - original draft, Writing - review \& editing}


\author[1,2,3]{Haoyu Wang}




\credit{Writing - review \& editing}


\author[1,2,3]{Mengwei Bao}



\credit{Writing - review \& editing}


\cortext[1]{Corresponding author.}



\begin{abstract}
With the rapid development of sensor and artificial intelligence (AI) technologies, the intelligent surface vehicles (ISVs) have gained increasing attention from both academia and industry. The corresponding intelligence, reliability and safety are highly dependent on the situational awareness in complex navigational environments. To obtain the high-quality perception capacity, we develop a vision-centered multi-sensor fusion system (named NavEYE) by taking full advantages of several different sensors, e.g., automatic identification system (AIS), radar and RGB camera, etc. In particular, we first propose a multi-constrained gated data association method (MCGA) to accurately match the low-temporal-resolution AIS and high-temporal-resolution radar data. Their fusion result is then obtained by selectively implementing the distance-aware adaptively-weighted fusion (DAWF) and timeliness decay-based stitching fusion (TDSF) methods, which could eliminate the uncertainty of loss of AIS or radar data in real-world sensing scenarios. Based on the visual object detection with high accuracy and robustness, we further associate and fuse the AIS, radar and visual data through joint constraints of normalized bearing and normalized distance features. According to the obtained fusion results, the comprehensive information (including visual appearance, identity, and distance, etc.) related to the ships of interest could be automatically obtained, beneficial for enhancing situational awareness and reducing collision risk for ISVs. The feasibility, robustness, usability, and effectiveness of our multi-sensor fusion method and situational awareness system have been demonstrated via numerous experiments on real-world sensing dataset from AIS, radar and camera, etc. The experimental results have illustrated the superior performance of our fusion method in terms of both quantitative and qualitative evaluations. In addition, our shipboard NavEYE system is capable of promoting navigational safety for ISVs in complex and dynamic environments.
\end{abstract}





\maketitle

\section {Introduction} \label{sec: introduction}

    Intelligent surface vehicles (ISVs), including unmanned surface vehicles (USVs) and autonomous surface vehicles (ASVs), represent a new generation of maritime transportation equipment \citep{liu2022intelligent, liu2025concepts}. They are driving the transition of maritime traffic from manual operation to intelligent assistance and fully autonomous navigation \citep{jovanovic2024review, xu2026challenges}. Maritime Situational Awareness, known as MSA, provides the foundation for path planning, collision avoidance decision making and cooperative control in ISVs by supporting comprehensive understanding and dynamic tracking of the surrounding environment. It is therefore a core prerequisite for intelligent navigation \citep{thombre2022sensors}. Accurate, comprehensive and stable perception of surrounding ships forms the lowest level of situational construction. It directly affects the reliability and safety of the upper level decision system. Therefore, research on adaptive and comprehensive methods for maritime ship object perception is important for the practical deployment of intelligent navigation technologies.

    Shipborne platforms are usually equipped with AIS, marine radar and visual imaging devices to acquire state information about surrounding ships. However, each sensor has limitations in sensing mechanism and data characteristics \citep{lee2023data}. AIS provides structured identity and motion information, but it has a low update rate, may suffer from signal loss and cannot perceive non cooperative objects \citep{harati2007ais}. Radar has strong detection capability for non cooperative objects, but it is easily affected by sea clutter and weather, and its accuracy decreases with distance \citep{bounaceur2022analysis, liu2023aioenet, liu2024real}. Visual devices provide rich semantic information, but they do not directly provide dynamic and static ship state information \citep{guo2023asynchronous}. A single sensor can hardly satisfy identity recognition, spatial accuracy and semantic representation at the same time \citep{lu2026cmivtp}. Therefore, the effective association and fusion of heterogeneous sensor data have become key technologies for improving ship object perception.

    Existing studies have achieved significant progress in multi-source sensor data association and fusion \citep{lu2026graph, zhang2025multi}. However, engineering applications for ship object perception still face multiple challenges \citep{qiao2021marine}. At the AIS–radar data association level, most existing methods rely on either single-feature constraints or computationally expensive graph-network-based approaches, making it difficult to adapt in real time to dynamic changes in object density and motion patterns. At the AIS–radar data fusion level, traditional weighted fusion methods generally assume constant sensor measurement errors and fail to fully consider the characteristic that radar observation accuracy varies dynamically with object distance. As a result, the fusion performance cannot maintain consistent accuracy across different distance ranges. In addition, abrupt track changes may occur during the dynamic switching between dual-source AIS-radar observation mode and single-source observation mode. Existing methods usually replace fusion results directly with a single-source observation. This direct replacement can lead to sudden changes in position, speed over ground (SOG), and course over ground (COG). Consequently, the continuity and smoothness of the generated trajectories are degraded. At the vision-AIS-radar data fusion level, existing methods mainly depend on single coordinate projection or bearing constraints. When multiple ships are located near the same bearing within the field of view, association ambiguity is likely to occur. Moreover, the distance uncertainty introduced by monocular vision further reduces fusion accuracy. From a system-level perspective, most existing studies remain at the algorithm validation stage and lack a complete shipborne engineering system.

    To address these issues, we propose a vision-AIS-radar multi-source data association and fusion framework, based on which the NavEYE situational awareness system is developed. First, a multi-constraint gating association (MCGA) method is adopted to achieve reliable matching between AIS-radar data, thereby establishing object correspondences for subsequent AIS-radar data fusion. Then, according to the availability of observation sources, the distance-aware adaptively-weighted fusion (DAWF) method and the timeliness decay-based stitching fusion (TDSF) method are selectively employed to obtain accurate and continuous AIS–radar fusion results. Subsequently, based on the fused AIS–radar data and visual detection results, the VARF method is used for vision-AIS-radar fusion to achieve cross-modal association between image-space objects and geospatial objects. Finally, the NavEYE system is developed based on the proposed framework, forming a vision-centered multi-sensor fusion situational awareness system for ISVs. The main contributions of this paper are as follows:

    \begin{itemize}
        \item The measurement errors of AIS and radar vary dynamically with object distance, and switching between single-source and dual-source observations can easily lead to trajectory discontinuities. To overcome these challenges, we propose a distance-aware adaptively-weighted (DAWF) method and a timeliness decay-based stitching fusion (TDSF) method. The proposed methods improve the accuracy, continuity, and robustness of fused trajectories.
        
        \item There is a cross-modal representation gap between visual detection objects and geospatial objects, which can easily lead to association ambiguities in multi-object scenarios. To overcome this challenge, we propose a vision-AIS-radar fusion (VARF) method based on dual constraints of bearing and distance. This method jointly models cross-modal normalized bearing and normalized distance features, thereby improving the accuracy and robustness of the association between visual objects and geospatial objects.
        
        \item To meet the practical navigation requirements of ISVs, this paper develops NavEYE, a vision-centered multi-sensor fusion situational awareness system. The system fully utilizes AIS, radar, vision, GNSS, and gyrocompass data, enabling end-to-end processing and visualization from raw observations to navigational assistance information.
        
        \item We construct the MAPFusion dataset for shipborne multi-sensor fusion. Experiments on MAPFusion show that the proposed method achieves competitive performance in AIS-radar data association, AIS-radar data fusion, and vision-AIS-radar data fusion tasks.
    \end{itemize}

    The remainder of this paper is organized as follows. Section~\ref{sec: related work} reviews related methods and existing situational awareness systems. Section~\ref{sec: method} presents the proposed shipborne vision-AIS-radar association and fusion framework. Section~\ref{sec: experiments and analysis} reports experiments and analyzes the results. Section~\ref{sec: NavEYE} introduces the NavEYE system. Section~\ref{sec: conclusions and discussion} concludes the paper and discusses future research directions.

\section{Related work} \label{sec: related work}

    Following the scope of this study, we review related work in terms of AIS-radar data association, AIS-radar data fusion, multi-sensor fusion methods, and shipborne situational awareness systems.

    \subsection{AIS-radar data association}
    
        AIS-radar data association is a classical problem in the field of ship object perception. Existing methods can be broadly grouped into statistical inference, fuzzy and grey association and deep learning. Statistical inference methods are represented by joint probabilistic data association (JPDA) \citep{fortmann1983sonar}, interacting multiple model probabilistic data association \citep{blom1988interacting}, and multiple hypothesis tracking (MHT) \citep{reid2003algorithm}. These methods use explicit probabilistic modeling to determine whether two observations come from the same source. However, they are sensitive to prior assumptions about noise distributions and motion models, and their performance decreases markedly in dense object and maneuvering scenarios. Fuzzy and grey association methods construct similarity measures using distance, bearing, SOG and COG \citep{yan2023association}. These methods reduce the influence of observation uncertainty, but their membership functions and weights depend on experience and have difficulty adapting to dynamic navigation environments. In recent years, deep learning methods based on graph neural networks and attention mechanisms \citep{yang2022multitarget} have shown stronger representation capability in multi source data association. However, their requirements for labeled data and computational resources are high, which makes them difficult to deploy in real time on shipborne platforms.

    \subsection{AIS-radar data fusion}
 
        Existing AIS-radar data fusion methods mostly employ weighted averaging based on fixed error variances \citep{wang2023intelligent, kazimierski2015radar}. They do not fully consider the different measurement characteristics of AIS and radar \citep{habtemariam2012measurement}. AIS position error mainly comes from the inherent noise of GNSS and is relatively stable \citep{fukuda2024study}. In contrast, shipborne radar position error is constrained by beam width and cross distance resolution. It increases significantly as object distance increases \citep{tian2009coordinate}. This distance dependent dynamic error means that the optimal fusion weights of the two sensors are not constant. A fusion strategy with constant weights can underestimate radar accuracy at short distance and overestimate radar contribution at long distance, thereby reducing the overall accuracy of fused tracks. In real navigation, object observation states can frequently switch between dual source AIS and radar cooperation and single source observation because of detection distance, sea clutter interference and communication link quality \citep{singh2020machine}. Existing methods often replace the fused result with a single data source during switching. This can cause abrupt changes in position, SOG and COG and damage the continuity and smoothness of track estimation.

    \subsection{Multi-sensor data fusion}

        Multi-sensor fusion has become a mainstream paradigm in intelligent perception because it can combine perception accuracy, semantic representation and all weather robustness. In autonomous driving, lidar and camera fusion methods \citep{liu2023bevfusion} unify point cloud geometric accuracy and image semantic information in bird eye view space. However, long distance lidar point clouds are sparse, and their effective operating distance is usually limited to 100 to 200 m. This does not satisfy maritime perception requirements, where ships are often several nautical miles away. Millimeter wave radar and camera fusion \citep{nabati2021centerfusion} shows stronger robustness in adverse weather. However, vehicle millimeter wave radar differs greatly from maritime marine radar in operating mechanism and point cloud density, so it is difficult to transfer directly to maritime scenarios. In the maritime field, existing fusion studies mostly focus on two modalities, vision and AIS \citep{yang2025tad,fang2026orientation}. \citet{lu2021fusion} estimated the relative bearing and distance of object ships from a monocular camera and matched them with AIS positions, but monocular estimation easily introduces uncertainty. \citet{gulsoylu2024image} proposed an vision and AIS association method based on homography transformation. This method relies only on position projection or a single bearing constraint. It is prone to association ambiguity in dense object scenarios and cannot perceive non-cooperative objects that do not carry AIS. Studies that jointly fuse vision, AIS and radar remain limited. \citet{huang2025surface} constructed a trimodal fusion dataset for inland waterways and proposed the multistage detection and tracking method MSTrack. However, their fusion strategies do not consider the spatial accuracy differences among sensors.

    \subsection{Shipborne situational awareness systems}

        Shipborne situational awareness systems are key carriers for engineering deployment of intelligent navigation technologies. Their core goal is to integrate multi source sensor observations in real time into a situational picture that can support navigation decisions. Early studies mainly focused on integrated display of AIS and radar by overlaying both types of information on Electronic Chart Display and Information Systems, known as ECDIS, to support navigation \citep{kazimierski2015radar}. With the development of computer vision and deep learning, optical sensors have gradually been included in shipborne perception systems. \citet{han2020autonomous} integrated radar, lidar, cameras and AIS in the ARAGON USV system. They validated the engineering feasibility of multi-sensor cooperative perception through field trials in multiple scenarios. This is a representative system level study in this field. \citet{prasad2017maritime} proposed an adaptive multi-sensor management architecture for low visibility conditions. It supports autonomous decision making by using cloud based cooperative fusion of AIS, vision sensors and meteorological data from own ship and other ships. For augmented reality assistance, \citet{fan2018research} proposed the shipborne assisted navigation system STDANS based on enhanced traffic environment perception. It combines three dimensional simulation with the real ship operating environment, improving crew perception of the water traffic environment. However, these systems share two main limitations. First, existing systems lack solutions to engineering issues such as the dynamic variation of AIS–radar observation accuracy with object distance and trajectory discontinuities caused by switching between single-source and dual-source observation modes. Second, most existing systems use vision sensors for object detection only. They do not deeply fuse visual detection results with AIS-radar geospatial objects. Therefore, they cannot provide unified fusion of object identity, motion state and visual appearance.

\section{Method} \label{sec: method}

\begin{figure*}[pos=!t]
        \centering
        \includegraphics[width=0.95\textwidth]{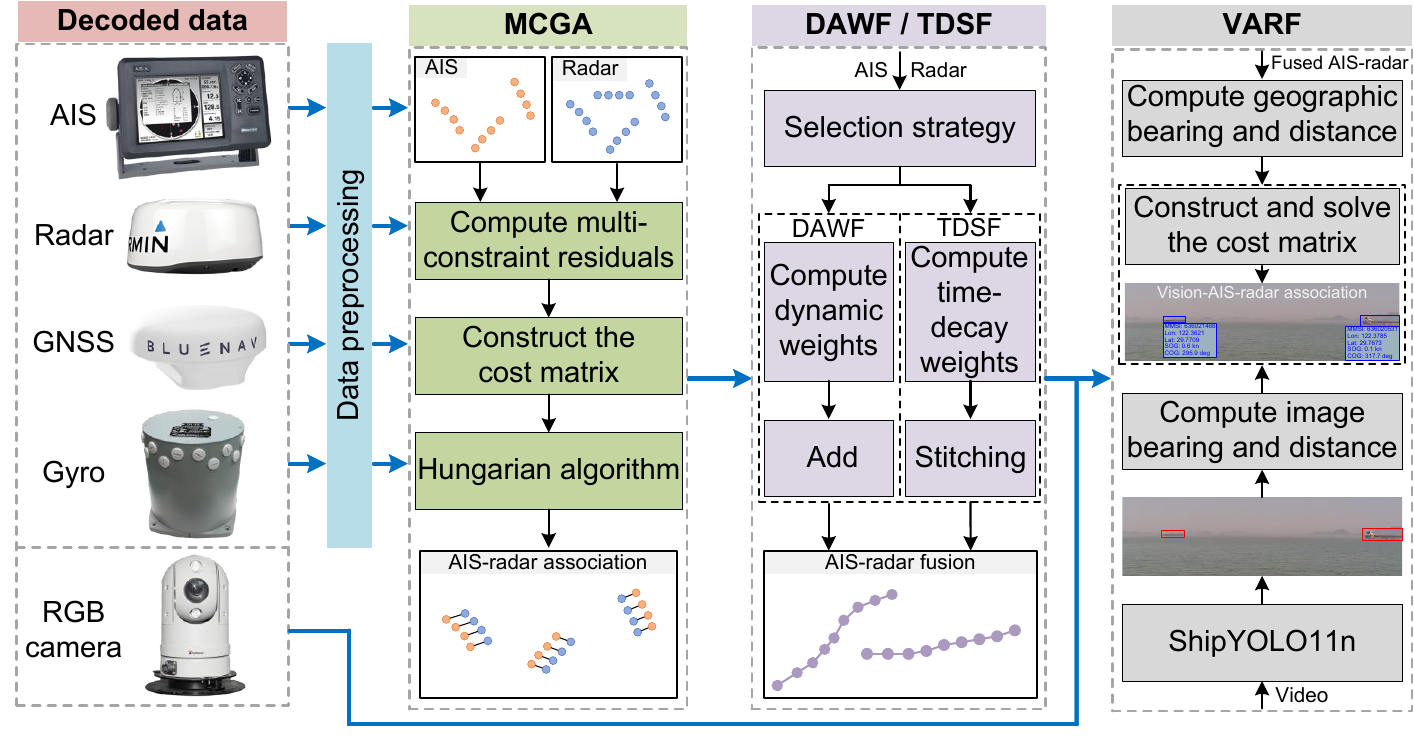}
        \caption{The framework of the proposed vision-centered multi-sensor data fusion. Gyro denotes the gyrocompass device. MCGA, DAWF/TDSF, and VARF denote the proposed AIS–radar data association, AIS–radar data fusion, and vision–AIS–radar data fusion methods, respectively. The selection strategy is used to select the appropriate fusion method according to the available observation sources. Fused AIS–radar denotes the fused AIS–radar data generated by the DAWF/TDSF module, which are used to compute the geographic bearing and distance in the VARF module.}
        \label{fig: method_overview}
    \end{figure*}

    To address the ship object perception requirements of ISVs, we propose a vision-centered vision-AIS-radar multi-sensor data association and fusion framework, as illustrated in Fig.~\ref{fig: method_overview}. First, multi-source sensor data from AIS, radar, GNSS, and the gyrocompass are decoded and preprocessed through threshold filtering, field-of-view filtering, key-field completeness checking, and value-range checking, thereby obtaining cleaned structured data. Subsequently, based on multidimensional state features, a multi-constraint gating method is adopted to achieve reliable association between AIS objects and radar objects. Then, by jointly considering object distance and the timeliness of observation information, an adaptive AIS-radar data fusion method is designed to improve the accuracy and continuity of track estimation. Finally, under dual constraints of bearing and distance, the fused geospatial objects are cross-modally associated with visual detection objects, enabling a unified representation of object identity, motion state, and visual appearance.

    \subsection{AIS-radar data association based on multi-constrained gating} \label{sec: aIS-radar data association}

        To achieve effective association between AIS and radar ship objects, we propose a multi-constrained gated data association method, named MCGA. This method uses object longitude, latitude, SOG and COG as association features. First, the Mahalanobis distance is employed to measure the discrepancy between AIS objects and radar objects, and candidate association pairs are screened through a gating mechanism. Subsequently, an association cost matrix is constructed, and the Hungarian algorithm is adopted to achieve optimal one to one association.

        Let the state vectors of the $i$-th AIS object and the $j$-th radar object at time $t$ be $\mathbf{\eta}_{a,t}^{i} = [lat_{a,t}^{i},lon_{a,t}^{i},sog_{a,t}^{i},cog_{a,t}^{i}]^{T}$ and $\mathbf{\eta}_{r,t}^{j} = [lat_{r,t}^{j},lon_{r,t}^{j},sog_{r,t}^{j},cog_{r,t}^{j}]^{T}$, respectively. where $lat$, $lon$, $sog$ and $cog$ denote latitude, longitude, SOG and COG. Considering the periodicity of the course angle, the course residual is defined in Eq. \eqref{eq: 1}. The state residual vector is then obtained by Eq. \eqref{eq: 2}. Since latitude, longitude, SOG, and COG have different physical meanings and units, directly measuring similarity based on raw residuals may be affected by differences in numerical scales, causing the association result to be dominated by dimensions with larger magnitudes or stronger error characteristics. Therefore, the Mahalanobis distance is adopted to measure the state residual between AIS and radar objects, presented in Eq. \eqref{eq: 3}. By normalizing and weighting the residuals according to the uncertainty of each state dimension, the Mahalanobis distance can reduce the influence of different units and measurement accuracies, thereby providing a more reasonable representation of the similarity between AIS and radar objects.

        \begin{equation}
            \Delta cog_{t}^{i,j} = \left[ cog_{a,t}^{i} - cog_{r,t}^{j} \right]_{-180^{\circ}}^{180^{\circ}}
            \label{eq: 1}
        \end{equation}

        \vspace{-4pt}

        \begin{equation}
            \begin{aligned}
            \Delta \eta_{t}^{i,j} = [&lat_{a,t}^{i}-lat_{r,t}^{j}, \ lon_{a,t}^{i}-lon_{r,t}^{j} \\[6pt] 
                                     & sog_{a,t}^{i}-sog_{r,t}^{j},\ \Delta cog_{t}^{i,j}]^{T}
            \end{aligned}
            \label{eq: 2}
        \end{equation}

        \vspace{-4pt}
        
        \begin{equation}
            D_{t}^{i,j} = \sqrt{({\Delta\mathbf{\eta}}_{t}^{i,j})^{T}\mathbf{S}^{-1}{\Delta\mathbf{\eta}}_{t}^{i,j}},
            \label{eq: 3}
        \end{equation}

        \noindent where $\left[\alpha\right]_{-180^{\circ}}^{180^{\circ}}=\left((\alpha+180^{\circ}) \bmod 360^{\circ}\right)-180^{\circ}$ denotes the angular normalization operator, and $\Delta cog_{t}^{i,j}$ represents the shortest COG difference between the $i$-th AIS object and the $j$-th radar object at time $t$. $\mathbf{S}=\operatorname{diag}(\sigma_{lat}^2,\sigma_{lon}^2,\sigma_{sog}^2,\sigma_{cog}^2)$ is the residual covariance matrix, in which $\sigma_{lat}$, $\sigma_{lon}$, $\sigma_{sog}$, and $\sigma_{cog}$ denote the standard deviations of latitude, longitude, SOG, and COG, respectively. $\mathbf{S}^{-1}$ denotes the inverse of $\mathbf{S}$.

        Before association decision making, a gating threshold $g$ is set to filter valid candidate association pairs. When $D_t^{i,j}\leq g$, the $i$-th AIS object and the $j$-th radar object are regarded as a valid candidate pair. Otherwise, they are regarded as unassociated. Based on the gating result, the association cost matrix at time $t$ is constructed as $\mathbf{C}_{t}=[c_t^{i,j}]_{N_a\times M_r}$, where $N_a$ and $M_r$ denote the numbers of AIS objects and radar objects. The matrix element is defined in Eq. \eqref{eq: 4}.

        \begin{equation}
            c_t^{i,j}=\begin{cases}
            D_t^{i,j}, & D_t^{i,j}\leq g  \\[5pt]
            M, & D_t^{i,j}>g,
            \end{cases}
            \label{eq: 4}
        \end{equation}
        
        \noindent where $M$ is a large penalty cost used to represent infeasible association. The Hungarian algorithm is used to solve the one to one optimal assignment problem so that the total association cost is minimized. AIS and radar object pairs that satisfy the gating condition and are successfully assigned are identified as the same ship.

    \subsection{Adaptive AIS-radar data fusion} \label{sec: adaptive AIS-radar data fusion}

\begin{figure*}[pos=!t]
            \centering
            \includegraphics[width=0.95\linewidth]{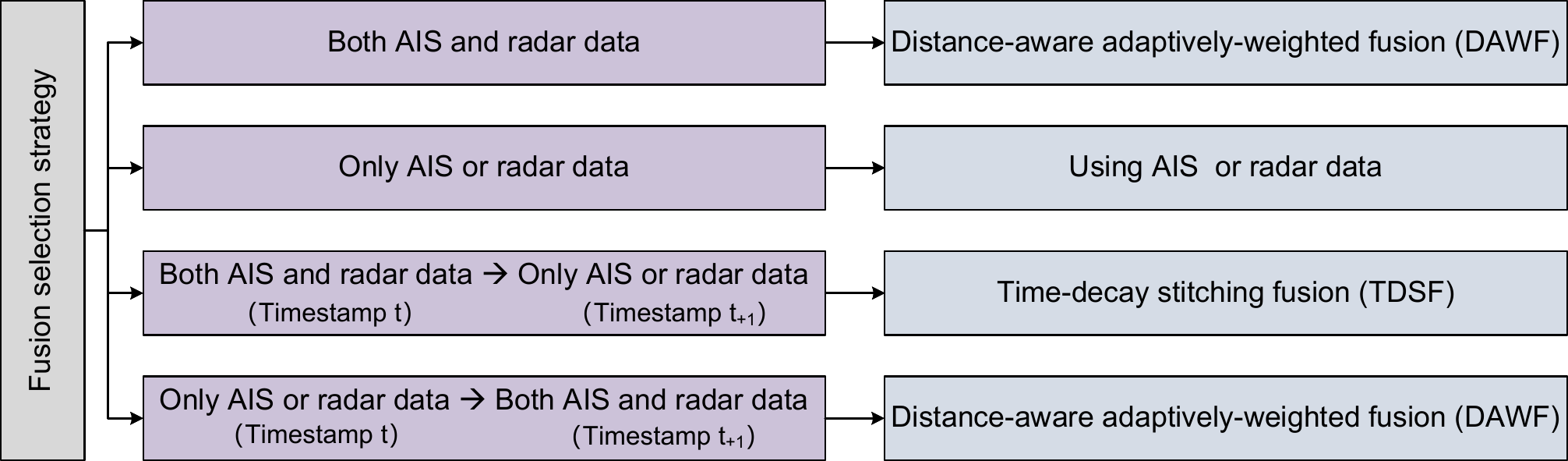}
            \caption{Adaptive AIS-radar data fusion strategy. DAWF is used when both AIS and radar data are available or when the observation state changes from single-source to dual-source; the available sensor trajectory is directly output under continuous single-source observations; and TDSF is applied when the observation state changes from dual-source to single-source.}
            \label{fig: fusion_selection_strategy}
        \end{figure*}

        After completing the association between AIS objects and radar objects, we propose an adaptive AIS–radar data fusion method to fully exploit the complementary information provided by the two sensors. This method comprehensively considers the influence of object distance on sensor observation accuracy, as well as the timeliness of historical fusion information during observation-source switching, As illustrated in Fig.~\ref{fig: fusion_selection_strategy}.
        
        \subsubsection{Distance-aware adaptively-weighted fusion}

            Considering dynamically varying sensor measurement errors with observation distance in real navigation environments, we propose a distance-aware adaptively-weighted fusion method, named DAWF. Based on the differences in the operating mechanisms of AIS and radar, dynamic error variance models related to object distance are established to achieve adaptive weighted fusion.

            The positional error of AIS data mainly originates from the inherent observation noise of the global navigation satellite system (GNSS), and its equivalent measurement error variance is given in Eq. \eqref{eq: 5}. Unlike AIS, shipborne radar, as an active microwave sensing device, exhibits position measurement accuracy that is significantly affected by object distance. As the object distance increases, radar position observation errors gradually increase due to the limitations of radar beamwidth and cross-range resolution. Therefore, the radar measurement error variance is modeled as a function of distance and given in Eq. \eqref{eq: 6}.
            
            \begin{equation}
                \sigma_a^2(d_t^i)=\sigma_{gnss}^2=1.0\times10^{-8}
                \label{eq: 5}
            \end{equation}

            \vspace{-4pt}
            
            \begin{equation}
                \sigma_r^2(d_t^i,\beta)=\sigma_{base}^2+\beta(d_t^i)^2=8.1\times10^{-5}+\beta (d_t^i)^2,
                \label{eq: 6}
            \end{equation}
            
            \noindent where $\sigma_{gnss}^2$ denotes the inherent GNSS baseline error variance. $d_t^i$ is the physical distance between own ship and the $i$-th object at time $t$. $\sigma_{base}^2$ denotes the inherent minimum measurement error variance of radar, and $\beta$ is the distance error coefficient.
    
            Based on the above dynamic error variances, fusion weights are assigned according to the variance minimization principle. The dynamic fusion weights of AIS and radar are computed by Eqs. \eqref{eq: 7}-\eqref{eq: 8}. The fused object state is obtained through weighted averaging by Eqs. \eqref{eq: 9}-\eqref{eq: 11}.

            \begin{equation}
                w_{a,t}^{i} = \frac{\sigma_r^2(d_t^i,\beta)}{\sigma_r^2(d_t^i,\beta)+\sigma_a^2(d_t^i)}
                \label{eq: 7}
            \end{equation}

            \vspace{-4pt}

            \begin{equation}
                w_{r,t}^{i} = \frac{\sigma_a^2(d_t^i)} {\sigma_r^2(d_t^i,\beta)+\sigma_a^2(d_t^i)}
                \label{eq: 8}
            \end{equation}

            \vspace{-4pt}

            \begin{equation}
                \begin{bmatrix}
                \vphantom{sog_{f,t}^{i}} lat_{f,t}^{i}  \\
                \vphantom{sog_{f,t}^{i}} lon_{f,t}^{i}  \\
                \vphantom{sog_{f,t}^{i}} sog_{f,t}^{i}
                \end{bmatrix}
                =
                w_{a,t}^{i}
                \begin{bmatrix}
                \vphantom{sog_{f,t}^{i}} lat_{a,t}^{i}  \\
                \vphantom{sog_{f,t}^{i}} lon_{a,t}^{i}  \\
                \vphantom{sog_{f,t}^{i}} sog_{a,t}^{i}
                \end{bmatrix}
                +
                w_{r,t}^{i}
                \begin{bmatrix}
                \vphantom{sog_{f,t}^{i}} lat_{r,t}^{i}  \\
                \vphantom{sog_{f,t}^{i}} lon_{r,t}^{i}  \\
                \vphantom{sog_{f,t}^{i}} sog_{r,t}^{i}
                \end{bmatrix}
                \label{eq: 9}
            \end{equation}

            \vspace{-4pt}
            
            \begin{equation}
                \Delta cog_{t}^{r,a} = \left[ cog_{r,t}^{i} - cog_{a,t}^{i} \right]_{-180^{\circ}}^{180^{\circ}}
                \label{eq: 10}
            \end{equation}

            \vspace{-4pt}
            
            \begin{equation}
                cog_{f,t}^{i} = \left[ cog_{a,t}^{i} + w_{r,t}^{i} \Delta cog_{t}^{r,a} \right]_{0^{\circ}}^{360^{\circ}}
                \label{eq: 11}
            \end{equation}
    
            \noindent Where $\left[\alpha\right]_{0^{\circ}}^{360^{\circ}}=(\alpha+360^{\circ}) \bmod 360^{\circ}$. $w_{a,t}^{i}$ and $w_{r,t}^{i}$ denote the fusion weights of the $i$-th AIS and radar objects at time $t$. $\mathbf{\eta}_{a,t}^{i}$ and $\mathbf{\eta}_{r,t}^{i}$ denote the object states provided by AIS and radar at time $t$. $\mathbf{\eta}_{f,t}^{i}$ denotes the fused state estimate of the $i$-th object at time $t$. Eqs. \eqref{eq: 5}-\eqref{eq: 8} show that when the object is close, radar has higher spatial resolution and the fused result relies more on radar observations. When the object distance increases, radar error rises quickly while AIS positioning error remains relatively stable. The fused result then relies more on AIS information. This method enables adaptive adjustment of fusion weights with object distance and improves track quality under dual source cooperative observation.

        \subsubsection{Timeliness decay-based stitching fusion}

\begin{figure}[pos=!t]
                \centering
                \includegraphics[width=0.90\columnwidth]{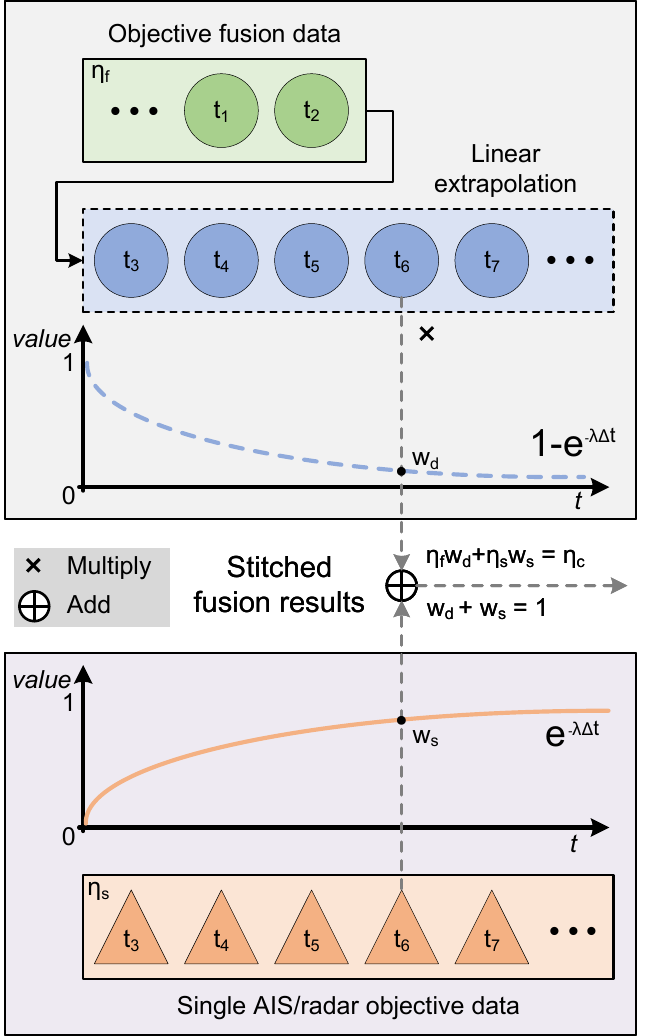}
                \caption{Timeliness decay-based stitching fusion method. It method can alleviate the problem of abrupt trajectory changes. $\eta_f$, $\eta_s$, and $\eta_c$ denote the extrapolated historical AIS--radar fused state, the current single-source observation state, and the stitched fusion result, respectively.}
                \label{fig: tdsf}
            \end{figure}

            In practical navigation environments, affected by factors such as limited detection range and sea clutter interference, the object observation state dynamically switches between the AIS–radar dual-source cooperative mode and the single-sensor observation mode. Directly using a single data source as the fusion result during mode switching may lead to abrupt changes in trajectory states such as position, SOG, and COG. To address this issue, we propose a timeliness decay-based stitching fusion method, named TDSF. By dynamically weighting historical fusion information and current single-source observations, the proposed method achieves smooth transitions during trajectory updates. The detailed process is illustrated in Fig.~\ref{fig: tdsf}.
    
            When a signal from one observation source is lost, the fused track generated before the loss can be regarded as a reliable prior estimate of the current object state because it integrates physical features from multiple sources. Based on the historical fused track, we use linear extrapolation to calculate the predicted state of the object at the current time $t$. Let the states of the $i$-th object at the previous two fused track times $t_{1}$ and $t_{2}$ be $\mathbf{\eta}_{f,t_{1}}^{i}=[lat_{f,t_{1}}^{i},lon_{f,t_{1}}^{i},sog_{f,t_{1}}^{i},cog_{f,t_{1}}^{i}]$ and {$\mathbf{\eta}_{f,t_{2}}^{i}=[lat_{f,t_{2}}^{i},lon_{f,t_{2}}^{i},sog_{f,t_{2}}^{i},cog_{f,t_{2}}^{i}]$, respectively. The predicted fused state $\widehat{\mathbf{\eta}}_{f,t}^{i}=[\widehat{lat}_{f,t}^{i},\widehat{lon}_{f,t}^{i},\widehat{sog}_{f,t}^{i},\widehat{cog}_{f,t}^{i}]$ at time $t$ is obtained by Eqs. \eqref{eq: 12}-\eqref{eq: 14}.

            \begin{equation}
                \begin{bmatrix}
                \vphantom{sog_{f,t}^{i}} \widehat{lat}_{f,t}^{i}  \\
                \vphantom{sog_{f,t}^{i}} \widehat{lon}_{f,t}^{i}  \\
                \vphantom{sog_{f,t}^{i}} \widehat{sog}_{f,t}^{i}
                \end{bmatrix}
                =
                \begin{bmatrix}
                \vphantom{sog_{f,t}^{i}} lat_{f,t_2}^{i}  \\
                \vphantom{sog_{f,t}^{i}} lon_{f,t_2}^{i}  \\
                \vphantom{sog_{f,t}^{i}} sog_{f,t_2}^{i}
                \end{bmatrix}
                \frac{t-t_1}{t_2-t_1}
                -
                \begin{bmatrix}
                \vphantom{sog_{f,t}^{i}} lat_{f,t_1}^{i}  \\
                \vphantom{sog_{f,t}^{i}} lon_{f,t_1}^{i}  \\
                \vphantom{sog_{f,t}^{i}} sog_{f,t_1}^{i}
                \end{bmatrix}
                \frac{t-t_2}{t_2-t_1}
                \label{eq: 12}
            \end{equation}

            \vspace{-4pt}
            
            \begin{equation}
                \Delta cog_{t_1,t_2}^{i} = \left[ cog_{f,t_2}^{i} - cog_{f,t_1}^{i} \right]_{-180^{\circ}}^{180^{\circ}}
                \label{eq: 13}
            \end{equation}

            \vspace{-4pt}
            
            \begin{equation}
                \widehat{cog}_{f,t}^{i} = \left[ cog_{f,t_2}^{i} + \frac{t-t_2}{t_2-t_1} \Delta cog_{t_1,t_2}^{i} \right]_{0^{\circ}}^{360^{\circ}}
                \label{eq: 14}
            \end{equation}

            To describe the reliability of historical fusion results after signal loss, we introduce a time-decay function associated with the delay $\Delta t$. A smaller $\Delta t$ indicates that the observation source was only recently lost, so the historical fusion result remains highly reliable. As $\Delta t$ increases, the validity of historical information gradually decreases. Accordingly, the weight of the historical fused track for the $i$-th object is defined in Eq. \eqref{eq: 15}, while the weight of the still-valid single-sensor observation is given in Eq. \eqref{eq: 16}.

            \begin{equation}
                w_d^i(\Delta t)=e^{-\lambda\Delta t}
                \label{eq: 15}
            \end{equation}

            \vspace{-5pt}
    
            \begin{equation}
                w_s^i(\Delta t)=1-w_d^i(\Delta t)=1-e^{-\lambda\Delta t},
                \label{eq: 16}
            \end{equation}

            \noindent where $\lambda$ is the decay constant that controls the weight decay rate. Finally, the stitched track point is generated by dynamically weighting the historical predicted state and the current single observation state, thereby achieving a smooth transition during track state switching, with the formulation given in Eqs. \eqref{eq: 17}-\eqref{eq: 19}.

            \begin{equation}
                \begin{bmatrix}
                \vphantom{\widehat{sog}_{f,\Delta t}^{i}} lat_{c,\Delta t}^{i}  \\
                \vphantom{\widehat{sog}_{f,\Delta t}^{i}} lon_{c,\Delta t}^{i}  \\
                \vphantom{\widehat{sog}_{f,\Delta t}^{i}} sog_{c,\Delta t}^{i}
                \end{bmatrix}
                =
                w_{d}^{i}(\Delta t)
                \begin{bmatrix}
                \vphantom{\widehat{sog}_{f,\Delta t}^{i}} \widehat{lat}_{f,\Delta t}^{i}  \\
                \vphantom{\widehat{sog}_{f,\Delta t}^{i}} \widehat{lon}_{f,\Delta t}^{i}  \\
                \vphantom{\widehat{sog}_{f,\Delta t}^{i}} \widehat{sog}_{f,\Delta t}^{i}
                \end{bmatrix}
                +
                w_{s}^{i}(\Delta t)
                \begin{bmatrix}
                \vphantom{\widehat{sog}_{f,\Delta t}^{i}} lat_{s,\Delta t}^{i}  \\
                \vphantom{\widehat{sog}_{f,\Delta t}^{i}} lon_{s,\Delta t}^{i}  \\
                \vphantom{\widehat{sog}_{f,\Delta t}^{i}} sog_{s,\Delta t}^{i}
                \end{bmatrix}
                \label{eq: 17}
            \end{equation}

            \vspace{-4pt}
            
            \begin{equation}
                \Delta cog_{\Delta t}^{s,f} = \left[ cog_{s,\Delta t}^{i} - \widehat{cog}_{f,\Delta t}^{i} \right]_{-180^{\circ}}^{180^{\circ}}
                \label{eq: 18}
            \end{equation}

            \vspace{-4pt}
            
            \begin{equation}
                cog_{c,\Delta t}^{i} = \left[ \widehat{cog}_{f,\Delta t}^{i} + w_{s}^{i}(\Delta t) \Delta cog_{\Delta t}^{s,f}\right]_{0^{\circ}}^{360^{\circ}}
                \label{eq: 19}
            \end{equation}
    
            \noindent where $\mathbf{\eta}_{f,\Delta t}^{i}$ denotes the fused predicted state of the $i$-th object at time delay $\Delta t$. $\mathbf{\eta}_{s,\Delta t}^{i}$ denotes the valid single source observation state of the $i$-th object at $\Delta t$. $\mathbf{\eta}_{c,\Delta t}^{i}$ denotes the state of the $i$-th object obtained after temporal stitching fusion at $\Delta t$. Eqs. \eqref{eq: 9} to \eqref{eq: 12} show that when observation source switching has just occurred, the system mainly inherits the previous fused track to suppress state jumps caused by hard switching. As time passes, the output gradually transitions to the real time single source observation, thereby ensuring timeliness and fidelity of the estimated result. This method effectively solves the track discontinuity problem during multi source observation state switching.

    \subsection{Vision-AIS-radar} data fusion based on bearing and distance \label{sec: vision–AIS–radar data fusion based on bearing and distance}

        To represent ship object identity, motion state and visual appearance in a unified manner, we propose a vision-AIS–radar data fusion method based on joint bearing and distance constraints, named VARF, as  illustrated in Fig.~\ref{fig: varf}. It uses AIS and radar fused tracks as geospatial observations and video detection objects as image space observations. By extracting relative bearing and relative distance features from the two types of observations, it constructs a cross space association cost matrix and uses the Hungarian algorithm to obtain globally optimal association. Finally, ship identity, position, SOG and COG information are mapped to video objects, achieving accurate fusion between visual objects and AIS and radar fused tracks.

        \begin{figure*}[pos=!t]
            \centering
            \includegraphics[width=0.90\textwidth]{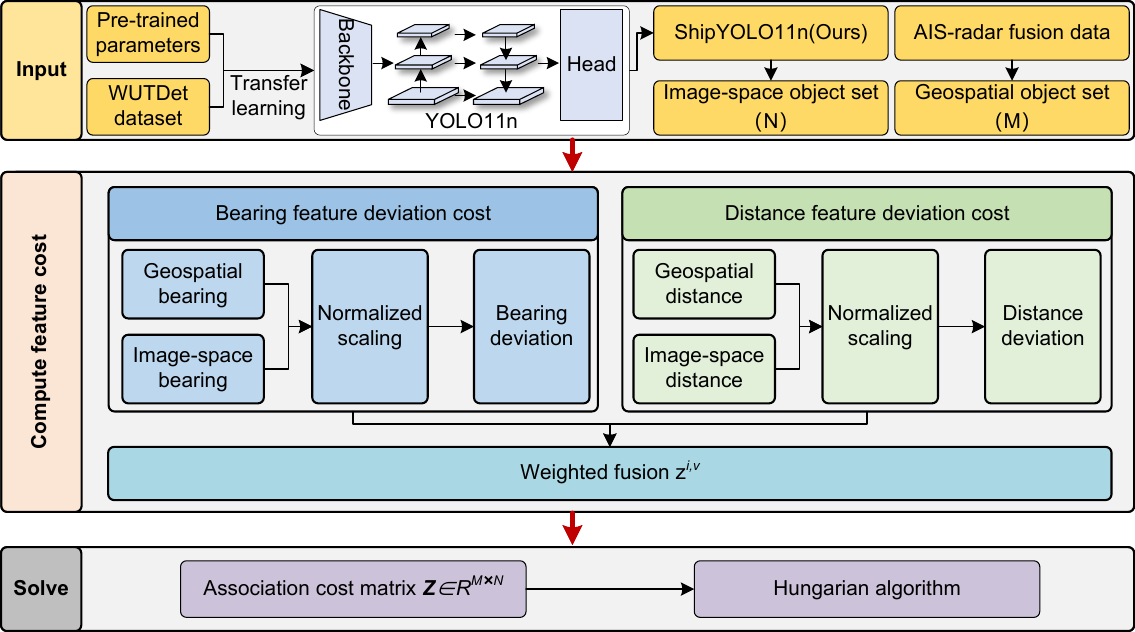}
            \caption{The illustration of the proposed vision–AIS–radar data fusion method. First, ShipYOLO11n trained via transfer learning extracts image-space and geospatial object sets from image data and AIS–radar fusion data, respectively. Then, bearing and distance feature deviation costs are computed through normalized and further fused to construct the association cost. Finally, the association cost matrix is generated and optimized using the Hungarian algorithm for cross-modal object fusion.}
            \label{fig: varf}
        \end{figure*}

        \subsubsection{Visual object detection}
        
            Real-time and accurate detection of ship objects \citep{liu2021enhanced, yang2025tad} in video is the basis of heterogeneous perception data fusion. Generic object detection models can be affected by changes in object scale, view angle and pose in complex waterway scenarios. This may lead to missed detections, false detections and unstable detection. We therefore use a strategy that combines a general detection framework with adaptation to a dedicated scenario dataset. This improves the model's ability to represent multi-scale and multi view ship objects. Considering the limited computing power and real time requirements of shipborne equipment, we select the lightweight YOLO11n \citep{jocher_Ultralytics_YOLO_2023} as the object detection model. We fine tune it on the WUTDet dataset \citep{liang2026wutdet} to obtain our ship object detection model, named ShipYOLO11n.

        \subsubsection{Geospatial and image-space object bearing} 
        
            To align and compare AIS and radar fused tracks with visual detection objects in a unified reference frame, the object bearing features in geospatial and image spaces must be computed separately and converted to relative bearing based on the COG of own ship. For an AIS and radar fused track object, the true north bearing between own ship and the object ship can be obtained from their latitude and longitude coordinates. Let the latitude and longitude of own ship at time $t$ be $(lat_t^{own},lon_t^{own})$, and let the latitude and longitude of the $i$-th object ship after AIS and radar fusion be $(lat_{f,t}^{i},lon_{f,t}^{i})$. The true bearing can be solved using the standard spherical geometry model in Eqs. \eqref{eq: 20}-\eqref{eq: 23}. Then, with the real time COG provided by the gyrocompass of own ship, the true north bearing can be converted to a relative bearing using Eq. \eqref{eq: 24}.

            \begin{equation}
                \Delta lon_{t}^{i} = lon_{f,t}^{i} - lon_{t}^{own}
                \label{eq: 20}
            \end{equation}

            \vspace{-4pt}
            
            \begin{equation}
                p_{t}^{i} =
                \cos(lat_{f,t}^{i}) \sin(\Delta lon_{t}^{i})
                \label{eq: 21}
            \end{equation}

            \vspace{-4pt}

            \begin{equation}
                \begin{aligned}
                    q_{t}^{i} &= \cos(lat_{t}^{own}) \sin(lat_{f,t}^{i})  \\[6pt] &\quad \times \sin(lat_{t}^{own}) \cos(lat_{f,t}^{i}) \cos(\Delta lon_{t}^{i})
                \end{aligned}
                \label{eq: 22}
            \end{equation}

            \vspace{-4pt}
            
            \begin{equation}
                tb_{t}^{i} =
                \left[
                \arctan2(p_{t}^{i},q_{t}^{i}) \frac{180}{\pi}
                \right]_{0^{\circ}}^{360^{\circ}}
                \label{eq: 23}
            \end{equation}

            \vspace{-4pt}
            
            \begin{equation}
                rb_t^i=\left[tb_t^i-cog_t^{own}\right]_{-180^{\circ}}^{180^{\circ}},
                \label{eq: 24}
            \end{equation}
            
            \noindent where $tb_t^i \in \left[0^{\circ}, 360^{\circ}\right)$ denotes the true bearing measured from true north, and $rb_t^i$ and $cog_t^{\mathrm{own}}$ denote the relative bearing and own-ship COG, respectively.
    
            Unlike geospatial objects, visual detection objects are located in the image plane coordinate system. The shipborne camera used in this study is fixed on the bow centerline. Image distortion has been corrected during preprocessing. Therefore, the image geometric centerline can be approximated as the COG reference line of own ship. Let the image width be $W$, and let the horizontal coordinate of the bottom midpoint of the $v$-th object box detected by ShipYOLO11n at time $t$ be $u_t^v$. Its pixel offset from the image centerline, $\Delta u_t^v$, is calculated by Eq. \eqref{eq: 25}. The equivalent focal length $f_o$ at the current resolution can be calculated using the camera horizontal field of view, $hfov$, by Eq. \eqref{eq: 26}. According to the perspective projection relationship, Eq. \eqref{eq: 27} maps the pixel offset to the relative bearing of the visual object.
            
            \begin{equation}
                \Delta u_t^v=\frac{W}{2}-u_t^v
                \label{eq: 25}
            \end{equation}

            \vspace{-4pt}
            
            \begin{equation}
                f_o=\frac{W}{2\tan(hfov/2)}
                \label{eq: 26}
            \end{equation}

            \vspace{-4pt}
            
            \begin{equation}
                \alpha_t^v=\arctan\left(\frac{\Delta u_t^v}{f_o}\right)\frac{180}{\pi},
                \label{eq: 27}
            \end{equation}
    
            \noindent where $\alpha_t^v$ denotes the deflection angle of the $v$-th visual object relative to the image centerline at time $t$.

        \subsubsection{Geospatial and image-space object distance} 
        
            Bearing alone cannot fully constrain cross modal association in multi object scenarios. This is especially true when several ships appear near the same bearing, where using only angular information can produce association ambiguity. Therefore, we introduce the relative distance between the object and own ship as a second constraint. For the $i$-th AIS and radar fused track object, the spherical distance between own ship and the object ship can be calculated from their latitude and longitude using the Haversine formula \citep{sinnott1984virtues}. Let the mean radius of the Earth be $R$. The absolute physical distance $D_{geo,t}^{i}$ between the object and own ship is obtained by Eqs. \eqref{eq: 28}-\eqref{eq: 30}. To compare it with the image space distance feature, we use the current maximum radar observation distance $D_{max}$ as the normalization reference. Eq. \eqref{eq: 31} normalizes the absolute distance to obtain $d_{geo,t}^{i}$.

            \begin{equation}
                A_t^i =
                \sin^2\left(\frac{lat_t^i - lat_t^{own}}{2}\right)
                \label{eq: 28}
            \end{equation}

            \vspace{-4pt}
            
            \begin{equation}
                B_t^i = \cos(lat_t^{own}) \cos(lat_t^i) \sin^2\left(\frac{lon_t^i - lon_t^{own}}{2}\right)
                \label{eq: 29}
            \end{equation}

            \vspace{-4pt}
            
            \begin{equation}
                D_{geo,t}^{i} =
                2R \cdot \arcsin \left(\sqrt{A_t^i + B_t^i}\right)
                \label{eq: 30}
            \end{equation}

            \vspace{-5pt}
            
            \begin{equation}
                d_{geo,t}^{i}=\frac{D_{geo,t}^{i}}{D_{max}},
                \label{eq: 31}
            \end{equation}
            
            \noindent where $d_{geo,t}^{i}\in[0,1]$. A smaller value indicates that the object is closer to own ship.
    
            For visual detection objects, direct monocular ranging introduces additional distance uncertainty and accumulated error. Since ship objects are located on the water surface, the vertical coordinate of the bottom midpoint of the detection box usually has a stable monotonic relationship with the actual object distance. We therefore use the vertical image position to construct the absolute visual distance feature. Let the image height be $H$, and let the vertical coordinate of the bottom midpoint of the $v$-th detection box be $y_{vis,t}^{v}$. The normalized visual distance feature is obtained by Eq. \eqref{eq: 32}.
    
            \begin{equation}
                d_{vis,t}^{v}=\frac{H-y_{vis,t}^{v}}{H},
                \label{eq: 32}
            \end{equation}
            
            \noindent where $d_{vis,t}^{v}\in[0,1]$. When the object is closer to the bottom of the image, $y_{vis,t}^{v}$ is larger and $d_{vis,t}^{v}$ is smaller. This corresponds to an object that is closer to own ship in the real scene.
    
            \subsubsection{Cost matrix and optimization} 
            
                We construct an association cost matrix with joint constraints from bearing deviation and distance deviation. First, because angles are periodic over $360^{\circ}$ and must be normalized, the bearing deviation between the $i$-th track object and the $v$-th visual object is calculated by Eq. \eqref{eq: 33}. Then, Eq. \eqref{eq: 34} computes the distance deviation between object $i$ and visual object $v$. Finally, the combined association cost is defined in Eq. \eqref{eq: 35}.
        
                \begin{equation}
                    \Delta\varphi_t^{i,v}=\frac{\min\left(|rb_t^i-\alpha_t^v|,360^{\circ}-|rb_t^i-\alpha_t^v|\right)}{\theta_{max}}
                    \label{eq: 33}
                \end{equation}

                \vspace{-5pt}
                
                \begin{equation}
                    \Delta d_t^{i,v}=|d_{geo,t}^{i}-d_{vis,t}^{v}|
                    \label{eq: 34}
                \end{equation}

                \vspace{-5pt}
                
                \begin{equation}
                    k_t^{i,v}=\omega_{\varphi}\Delta\varphi_t^{i,v}+\omega_d\Delta d_t^{i,v} \ \mathrm{with}\ \omega_{\varphi}+\omega_d=1,
                    \label{eq: 35}
                \end{equation}
        
                \noindent where $\theta_{max}$ is the maximum angular error threshold allowed by the system. $\omega_{\varphi}$ and $\omega_d$ denote the weights of the bearing feature and the distance feature. Dynamic adjustment of these two weights allows the method to adapt flexibly to different ship navigation scenarios.
        
                AIS, radar and vision sensors can have missed detections, false detections and inconsistent object numbers in real observations. Direct global assignment may therefore cause incorrect associations between objects with large spatial differences. To address this, we introduce an association threshold $\gamma$ and remove candidate association pairs whose costs exceed a reasonable distance. The association cost matrix at time $t$ is $\mathbf{Z}_{t}=[z_t^{i,v}]_{N_a\times M_r}$, and its elements are defined in Eq. \eqref{eq: 36}.
        
                \begin{equation}
                    z_t^{i,v}=\begin{cases}
                    k_t^{i,v}, & k_t^{i,v}\leq \gamma  \\[5pt]
                    M, & k_t^{i,v}>\gamma,
                    \end{cases}
                    \label{eq: 36}
                \end{equation}
                
                \noindent where $z_t^{i,v}$ denotes the combined association cost between the $i$-th AIS and radar fused track and the $v$-th visual object. The Hungarian algorithm is then used to solve the association cost matrix $\mathbf{Z}_{t}$ and obtain the globally optimal correspondence between visual objects and AIS and radar fused tracks.

\section{Experiments and analysis} \label{sec: experiments and analysis}

    \subsection{Experimental data}

        \begin{figure*}[pos=!t]
            \centering
            \includegraphics[width=0.90\textwidth]{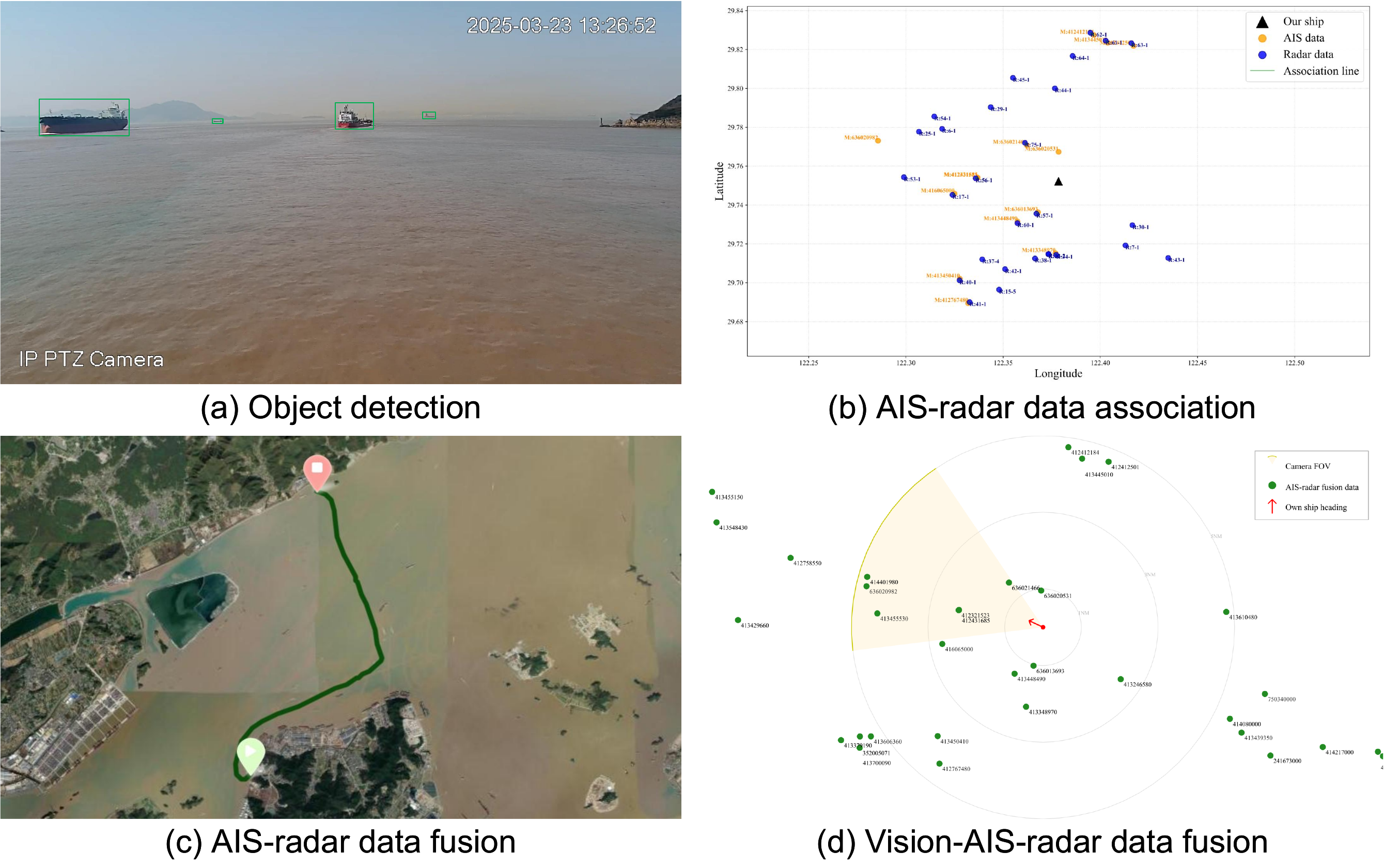}
            \caption{The examples of MAPFusion dataset annotations. From top left to bottom right: (a) object bounding boxes in video frames, (b) ground truth AIS-radar data association, (c) ground truth AIS-radar fused trajectory, and (d) ground truth vision-AIS-radar data fusion.}
            \label{fig: label}
        \end{figure*}

        We collected real shipborne multi-source data and constructed a multi-sensor fusion dataset (named MAPFusion). The dataset contains time-synchronized video streams, AIS, radar, GNSS, and gyrocompass data, which can be used for research on multimodal fusion tasks.

        \subsubsection{Data collection} 

            We employed multiple onboard sensors for data acquisition, including a visible-light camera, AIS receiver, radar, GNSS receiver, and gyrocompass. The visible-light camera captured images with a resolution of $1920 \times 1080$. It was installed at a height of $12.5 \mathrm{m}$ with a pitch angle of $10.0^\circ$, and its horizontal and vertical fields of view were $62.82^\circ$ and $36.70^\circ$, respectively. The radar was operated with the ARPA function in manual plotting mode, and the radar display range was switched according to the actual navigation conditions and object distribution. The experimental route started from Dinghai South Port in Zhoushan, Zhejiang (122.07$^\circ$E, 30.01$^\circ$N), passed through the Mazhi Channel, Taohua Channel, and Aoshan Channel, and reached the southern waters of Zhujiajian Island (122.42$^\circ$E, 29.82$^\circ$N) before returning. The entire voyage covered approximately 0.24$^\circ$ in latitude and 0.35$^\circ$ in longitude, encompassing typical navigation environments such as open waters, narrow waterways, and ports/anchorages.

            \begin{table*}[pos=!t]
                \centering
                \caption{The details of the MAPFusion dataset. It summarizes the video duration, number of video frames, number of AIS tracks, number of radar tracks, and object distribution characteristics across different scenarios.}
                \renewcommand{\arraystretch}{1.35}
                \setlength{\tabcolsep}{8pt}
                \label{tab: MAPFusion}
                \begin{tabular}{ l c c c c p{5cm} }
                    \toprule
                    Scenario          & Duration (min)  & Video frames  & AIS tracks  & Radar tracks  & Object distribution characteristics                                    \\
                    \midrule
                    Open waters       & 50              & 3000          & 1260        & 771           & Low object density, \newline few background interferences              \\
                    Narrow waterways  & 30              & 1800          & 857         & 534           & Medium object density, \newline longitudinal ship crossings occur      \\
                    Ports/anchorages  & 55              & 3300          & 1285        & 901           & High object density, \newline frequent occlusions between ships        \\
                    Total             & 135             & 8100          & 3402        & 2206          & -                                                                      \\
                    \bottomrule
                \end{tabular}
            \end{table*}
    
        \subsubsection{Dataset annotation}

            Using the video frame timestamp as the global temporal reference, we temporally aligned the raw multi-source sensor data through a kinematic prediction method, thereby obtaining synchronized multi-source observations at a 1 s interval. Each synchronized sample contains a video frame, AIS data, radar data, GNSS data, and gyrocompass data. The annotation process of the MAPFusion dataset consists of four parts: AIS–radar association labeling, ground truth collection for AIS–radar data fusion, ship object bounding box annotation, and vision–AIS–radar association labeling. Fig.~\ref{fig: label} (a) shows an example of object detection bounding box annotation, where ship objects in video frames are annotated using the labelImg\footnote{https://github.com/HumanSignal/labelImg} tool. Fig.~\ref{fig: label} (b) presents an example of AIS–radar association labeling. A dynamic geographic visualization platform was built based on Folium to spatially overlay the synchronized AIS and radar objects, and the one-to-one correspondence between AIS and radar objects was determined by manually comparing object positions and motion consistency. Fig.~\ref{fig: label} (c) shows the ground truth collection for AIS-radar data fusion. The roll-on/roll-off ship Zhoudou 16 (MMSI: 412439450, length 62 m, beam 14 m) is selected as the reference object because of its fixed route and regular navigation behavior. Its ground truth trajectory was provided by a GoPro Hero 6 GPS logger installed on the object ship. Fig.~\ref{fig: label} (d) shows an example of vision-AIS-radar association labeling. The positions of surrounding ships were projected onto a polar-coordinate map using the own ship’s GNSS position and gyrocompass COG as references, and were then compared and verified against the camera images to determine the matching relationship between object detection boxes and AIS-radar fused data. We strictly controlled the annotation quality. The GoPro Hero 6 GPS used for collecting the ground truth for AIS–radar data fusion has a horizontal positioning error of approximately 3–5 m. The AIS–radar association labels, object detection bounding boxes, and vision-AIS-radar association labels were all manually reviewed to ensure annotation accuracy.

        \subsubsection{Dataset statistics}
        
            Based on navigation environment complexity and traffic density, the MAPFusion dataset is categorized into three representative scenarios: open waters, narrow waterways, and ports/anchorages. After rigorous manual verification, the MAPFusion dataset contains 135 minutes of synchronized multi-sensor data. A total of 8,100 image frames are obtained by sampling at 1 FPS (with the original video frame rate of 25 FPS). The dataset includes 3,402 AIS tracks and 2,206 radar tracks, with detailed statistics summarized in Table~\ref{tab: MAPFusion}. On this basis, 1,783 images are annotated, resulting in 75,781 AIS–radar association pairs and 5,541 vision–AIS–radar association pairs for experimental evaluation. All images were used for testing.

    \subsection{Implementation details}

        All object detection models were implemented using PyTorch and trained on an Ubuntu-based high-performance computing workstation equipped with a 24~GB NVIDIA GeForce RTX 4090 GPU. During training, the input image size of ShipYOLO11n was set to \(1920 \times 1080\), while that of the other models was set to \(640 \times 640\). All models were trained for 300 epochs. All other experiments are conducted on windows using Python. The hardware platform includes an Intel Core i5 12490F 3.00 GHz CPU, 32 GB memory and a 16 GB NVIDIA GeForce RTX 4060 Ti GPU. In Eq. \eqref{eq: 3}, the parameters $\sigma_{lat}$, $\sigma_{lon}$, $\sigma_{sog}$, and $\sigma_{sog}$ are set to 0.005, 0.005, 3.0, and 3.0, respectively. In Eq. \eqref{eq: 4}, the parameter $g$ is set to 3.0. In Eqs. \eqref{eq: 15}-\eqref{eq: 16}, the parameter $\lambda$ is set to 0.5. In Eqs. \eqref{eq: 35}-\eqref{eq: 36}, the parameters $\omega_\phi$, $\omega_d$, and $\gamma$ are set according to different scenarios. For the open water scenario, $\omega_\phi$, $\omega_d$, and $\gamma$ are set to 0.65, 0.35, and 0.85, respectively. For the narrow-waterway scenario, $\omega_\phi$, $\omega_d$, and $\gamma$ are set to 0.60, 0.40, and 0.50, respectively. For the port/anchorage scenario, $\omega_\phi$, $\omega_d$, and $\gamma$ are set to 0.65, 0.35, and 0.80, respectively.

    \subsection{Evaluation metrics}

        For AIS and radar data association experiments, Precision, Recall and $F_1$-Score are used as evaluation metrics. They are calculated by Eqs. \eqref{eq: 37}-\eqref{eq: 39}. Precision measures the accuracy of predicted association relationships. It is the proportion of true positive samples among all samples predicted as positive. Recall measures the coverage of true association relationships. It is the proportion of correctly detected samples among all true positive samples. $F_1$-Score is the harmonic mean of precision and recall and reflects the balance between precision and recall capability.

        \begin{equation}
            Precision =
            \frac{TP}{TP+FP}
            \label{eq: 37}
        \end{equation}

        \vspace{-4pt}
        
        \begin{equation}
            Recall =
            \frac{TP}{TP+FN}
            \label{eq: 38}
        \end{equation}

        \vspace{-4pt}
        
        \begin{equation}
            F_1 =
            2\frac{Precision \cdot Recall}{Precision + Recall}
            \label{eq: 39}
        \end{equation}
        
        \noindent where $TP$ denotes the number of correctly associated positive samples. $FP$ denotes the number of incorrectly associated negative samples. $FN$ denotes the number of true association samples not identified by the model.

        For AIS and radar data fusion experiments, DTW, MAE and RMSE are used as evaluation metrics and are calculated by Eqs. \eqref{eq: 40}-\eqref{eq: 42}. DTW measures the similarity between the AIS and radar fused track and the true track. MAE and RMSE measure the deviation between the fused result and the ground truth.

        \begin{table*}[pos=!t]
                \centering
                \caption{Performance comparison of different AIS-radar data association methods on the MAPFusion. $\uparrow$ indicates that higher values are better. Bold values indicate the best results.}
                \renewcommand{\arraystretch}{1.35}
                \setlength{\tabcolsep}{12pt}
                \label{tab: ais-radar-association}
                \begin{tabular}{ l l c c c }
                    \toprule
                    Scenario                            & Method                            & $Precision(\%)$ $\uparrow$            & $Recall(\%)$ $\uparrow$         & $F_1$-Score $\uparrow$         \\
                    \midrule
                    \multirow{4}{*}{Open water}         & JPDA \citep{fortmann1983sonar}    & 67.88                                 & \textbf{87.24}                  & 0.7621                         \\
                                                        & MHT \citep{reid2003algorithm}     & 54.90                                 & 81.07                           & 0.6537                         \\
                                                        & GRA \citep{ju1982control}         & 81.94                                 & 83.41                           & 0.8258                         \\
                                                        & MCGA(Ours)                        & \textbf{84.61}                        & 83.35                           & \textbf{0.8381}                \\
                    \midrule
        
                    \multirow{4}{*}{Narrow channel}     & JPDA \citep{fortmann1983sonar}    & 53.71                                 & \textbf{88.47}                  & 0.6684                         \\
                                                        & MHT \citep{reid2003algorithm}     & 33.13                                 & 68.32                           & 0.4462                         \\
                                                        & GRA \citep{ju1982control}         & 82.07                                 & 85.35                           & 0.8367                         \\
                                                        & MCGA(Ours)                        & \textbf{83.62}                        & 86.88                           & \textbf{0.8520}                \\
                   \midrule
                   \multirow{4}{*}{Port/Anchorage}      & JPDA \citep{fortmann1983sonar}    & 50.92                                 & \textbf{80.66}                  & 0.6207                         \\
                                                        & MHT \citep{reid2003algorithm}     & 31.93                                 & 66.84                           & 0.4295                         \\
                                                        & GRA \citep{ju1982control}         & 60.40                                 & 73.05                           & 0.6613                         \\
                                                        & MCGA(Ours)                        & \textbf{82.02}                        & 74.61                           & \textbf{0.7810}                \\
                    \midrule
                    \multirow{4}{*}{All}                & JPDA \citep{fortmann1983sonar}    & 57.87                                 & \textbf{84.39}                  & 0.6824                         \\
                                                        & MHT \citep{reid2003algorithm}     & 40.95                                 & 72.54                           & 0.5183                         \\
                                                        & GRA \citep{ju1982control}         & 81.88                                 & 80.21                           & 0.8093                         \\
                                                        & MCGA(Ours)                        & \textbf{83.26}                        & 79.86                           & \textbf{0.8139}                \\
                    \bottomrule
                \end{tabular}
            \end{table*}

        \begin{equation}
            \begin{aligned}
            D(i,j) &= d(\widehat{\Theta}_i,\Theta_j) + \min\{D(i-1,j),  \\[6pt] &\qquad D(i,j-1),D(i-1,j-1)\}
            \end{aligned}
            \label{eq: 40}
        \end{equation}

        \vspace{-4pt}
        
        \begin{equation}
            MAE = \frac{1}{N}\sum_{i=1}^{N}|\Theta_i-\widehat{\Theta}_i|
            \label{eq: 41}
        \end{equation}

        \vspace{-4pt}
        
        \begin{equation}
            RMSE = \sqrt{\frac{1}{N}\sum_{i=1}^{N}(\Theta_i-\widehat{\Theta}_i)^2}
            \label{eq: 42}
        \end{equation}
        
        \noindent where $D(i,j)$ is the optimal warping cost of the cumulative distance matrix at $(i,j)$. $d(\widehat{\Theta}_i,\Theta_j)$ is the spatial distance between the fused object at time $i$ and the true object at time $j$. $N$ is the total number of samples. $\Theta_i$ is the true value of the $i$-th sample, and $\widehat{\Theta}_i$ is the predicted value of the $i$-th sample.

        For vision-AIS–radar data fusion experiments, $mAP_{50}$, $mAP_{50:95}$}, FPS and $Acc_{association}$ are used as evaluation metrics. The metrics mAP and $Acc_{association}$ are defined in Eqs. \eqref{eq: 43}-\eqref{eq: 44}. $mAP_{50}$ reflects the average detection accuracy when the IoU threshold is 0.50. $mAP_{50:95}$ is the mean average precision when the IoU threshold changes from 0.50 to 0.95 with a step of 0.05. It evaluates model robustness under different localization accuracy requirements. FPS denotes the number of images processed per second and measures computational efficiency. $Acc_{association}$ measures the association accuracy between visual object bounding boxes and position data.

        \begin{equation}
            mAP =\int_0^1 P(R)dR
            \label{eq: 43}
        \end{equation}

        \vspace{-4pt}
        
        \begin{equation}
            Acc_{association} = \frac{N_{correct}}{N_{total}}\times100\%
            \label{eq: 44}
        \end{equation}
        
        \noindent where $N_{correct}$ denotes the number of heterogeneous sensor object pairs correctly associated by the algorithm. $N_{total}$ denotes the number of manually annotated ground truth associated object pairs.

    \subsection{Results and discussion}

            \subsubsection{Performance of different AIS-radar data association methods}

\begin{figure*}[pos=!t]
                \centering
                \includegraphics[width=0.90\textwidth]{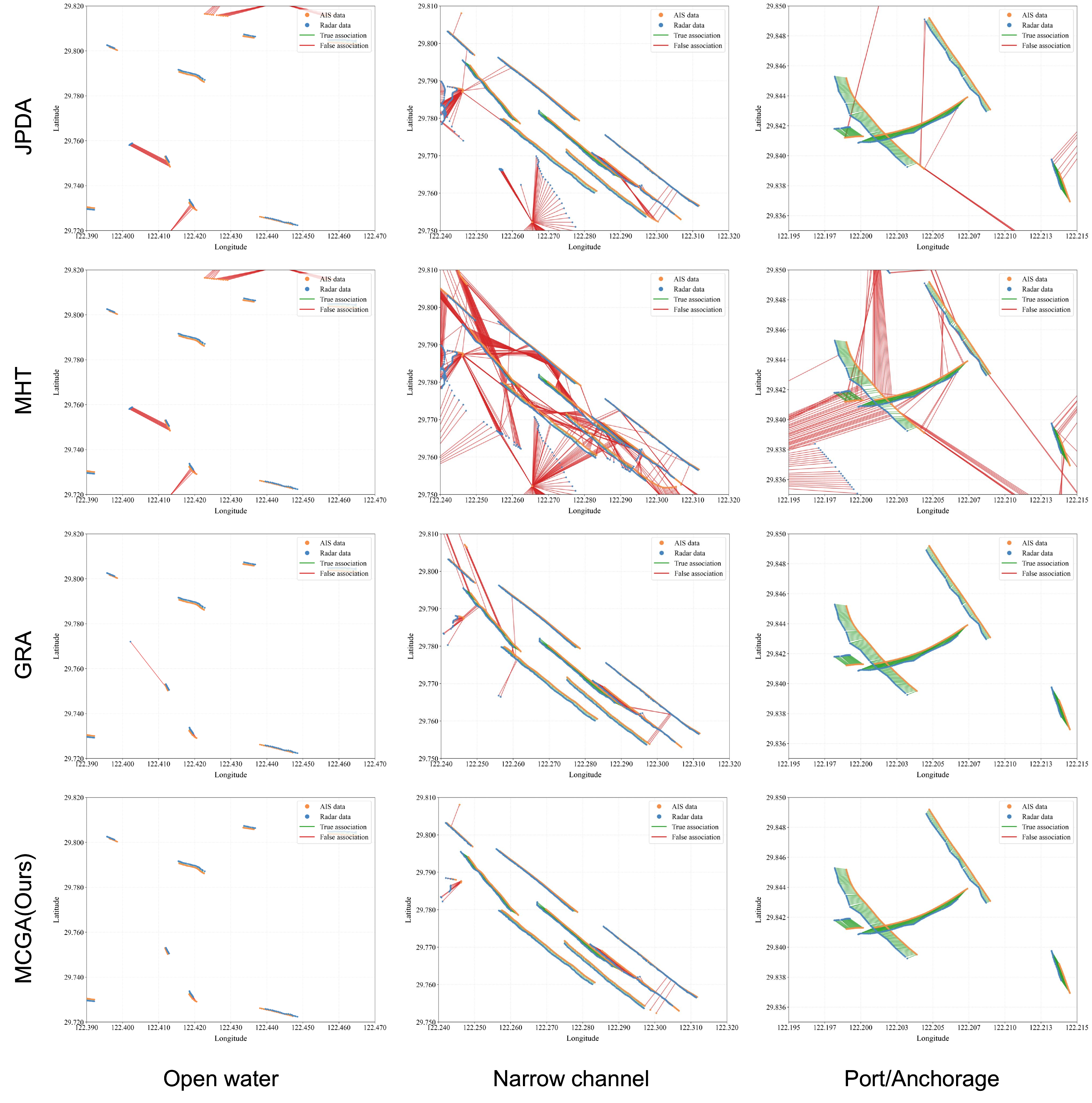}
                \caption{Visualization of AIS–radar data association results for different methods. MCGA (ours) exhibits fewer false associations across all scenarios, indicating better association stability and accuracy in complex maritime environments.}
                \label{fig: ais-radar-association}
            \end{figure*}

            To evaluate the effectiveness of the proposed MCGA method, we compare it with JPDA, MHT, and GRA. The experimental results are shown in Table~\ref{tab: ais-radar-association}. Across all scenarios, MCGA achieves the best performance in terms of precision (83.26\%) and $F_1$-score (0.8139), improving the $F_1$-score by 0.0046, 0.1315, and 0.2956 over GRA, JPDA, and MHT, respectively. In open waters, where objects are sparse and motion patterns are relatively simple, the performance gap among the four methods is small. In narrow waterways and port/anchorage scenarios, where object density increases and frequent longitudinal crossings and occlusions occur, the precision of JPDA drops to 50.92\%, while MHT further degrades to 31.93\%, and GRA decreases from 82.07\% to 60.40\%. In contrast, MCGA maintains stable precision ranging from 82.02\% to 84.61\% across all scenarios, with the fluctuation limited to only 2.59\%, demonstrating strong robustness to varying scene complexity. The visualization results in Fig.~\ref{fig: ais-radar-association} further support the above analysis. In the Narrow channel and port/anchorage scenarios, JPDA and MHT produce more one-to-many and crossing association errors, while GRA still shows a few incorrect links between adjacent objects. In contrast, the association lines generated by MCGA are highly consistent with the manual annotations.

            \begin{table*}[pos=!t]
                \centering
                \caption{Performance comparison of different AIS-radar data fusion methods on the MAPFusion using both AIS and radar data. RGF and AGF denote using radar-sensed object positions and AIS-sensed object positions as the fusion results, respectively. WAF denotes the weighted average fusion method. KF and EKF refer to the kalman filter and extended kalman filter, respectively. $\downarrow$ indicates that lower values correspond to better performance.}
                \renewcommand{\arraystretch}{1.35}
                \setlength{\tabcolsep}{15pt}
                \label{tab: weighted_fusion}
                \begin{tabular}{ l c c c c c c }
                    \toprule
                    Method                                & \multicolumn{2}{c}{Location (Lon., Lat.)}                     & \multicolumn{2}{c}{SOG}                            & \multicolumn{2}{c}{COG}                       \\
                                                          \cmidrule(lr){2-3}                                              \cmidrule(lr){4-5}                                   \cmidrule(lr){6-7}
                                                          & DTW $\downarrow$              & RMSE $\downarrow$             & MAE $\downarrow$     & RMSE $\downarrow$           & MAE $\downarrow$     & RMSE $\downarrow$      \\
                    \midrule
                    RGF                                   & 177.50 $\pm$ 114.28           & 214.88                        & 1.73                 & 2.48                        & 11.58                & 14.40                  \\
                    AGF                                   & 77.89 $\pm$ 38.84             & 87.04                         & 0.71                 & 0.83                        & 15.20                & 34.12                  \\
                    WAF                                   & 87.09 $\pm$ 47.42             & 100.14                        & 0.94                 & 1.25                        & 10.87                & 19.31                  \\
                    KF \citep{kalman1960new}              & 68.00 $\pm$ 36.67             & 77.53                         & 0.92                 & 1.43                        & 11.60                & 26.84                  \\
                    EKF \citep{schmidt1966application}    & 67.36 $\pm$ 35.91             & 76.47                         & 0.71                 & 0.89                        & 10.49                & 23.89                  \\
                    DAWF (Ours)                           & \textbf{58.97 $\pm$ 36.95}    & \textbf{69.59}                & \textbf{0.64}        & \textbf{0.75}               & \textbf{9.86}        & \textbf{12.63}         \\
                    \bottomrule
                \end{tabular}
            \end{table*}

        \subsubsection{Performance of different AIS-radar data fusion methods}

\begin{figure*}[pos=!t]
                \centering
                \includegraphics[width=0.93\textwidth]{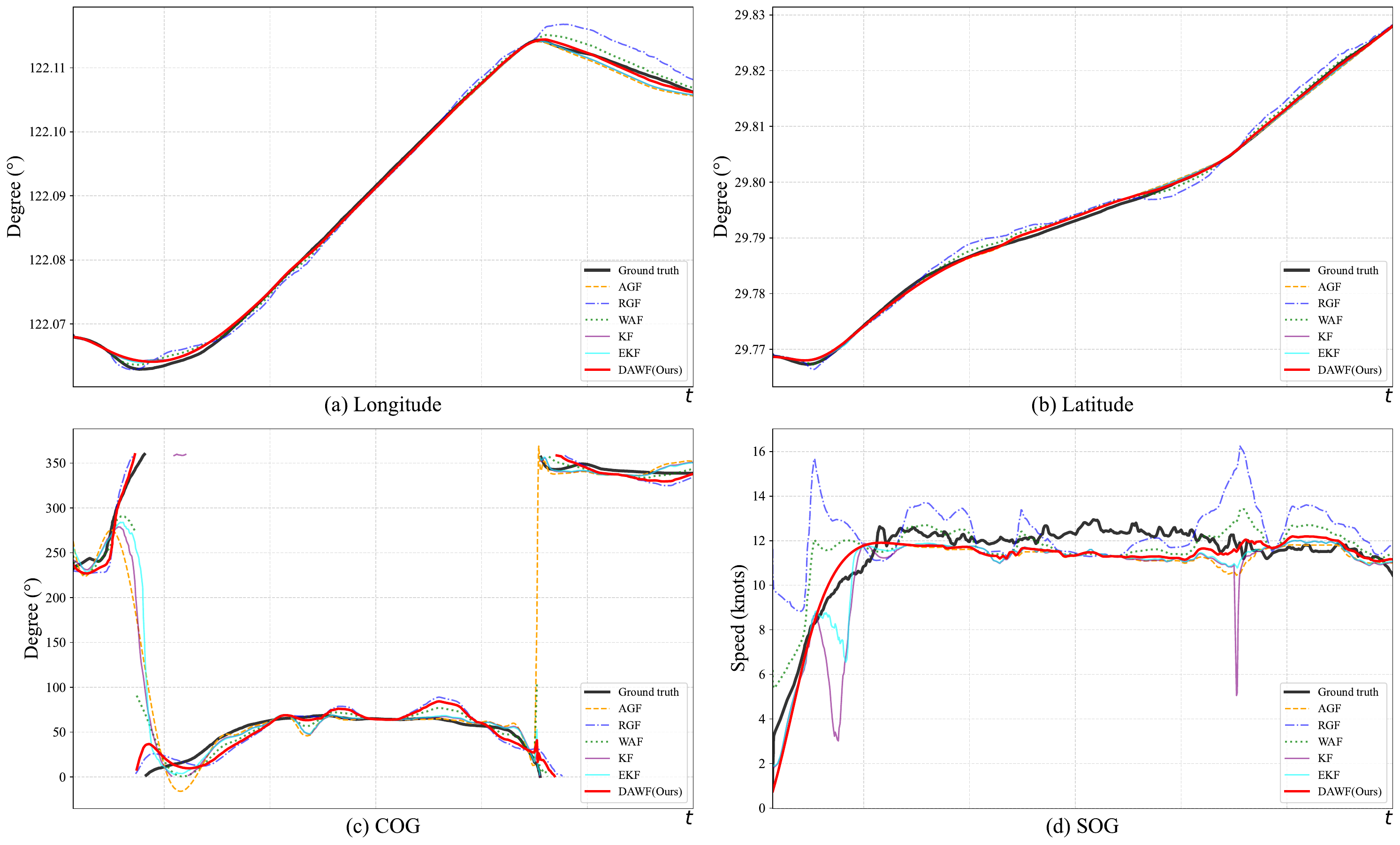}
                \caption{Visualization of AIS–radar data fusion results for different methods with both AIS and radar data available.}
                \label{fig: weighted_fusion}
            \end{figure*}

            To verify the accuracy advantage of DAWF under dual-source cooperative observation conditions, Table~\ref{tab: weighted_fusion} compares it with RGF, AGF, WAF, KF \citep{kalman1960new}, and EKF \citep{schmidt1966application}. In terms of location (Lon., Lat.), DAWF achieves the lowest DTW (58.97) and RMSE (69.59), reducing them by 12.5\% and 9.0\% compared with the second-best EKF, and by 32.3\% and 30.5\% compared with the fixed-weight fusion method WAF, respectively. For SOG estimation, DAWF also achieves the best MAE (0.64) and RMSE (0.75), with reductions of approximately 9.9\% and 9.6\% over the second-best results, respectively. For COG estimation, DAWF also obtains the best MAE ($9.86^\circ$) and RMSE ($12.63^\circ$), reducing them by approximately 6.0\% and 12.3\% compared with the second-best results, respectively. Overall, DAWF achieves the best performance in location, SOG, and COG estimation, indicating that it can effectively improve the fusion accuracy under AIS-radar dual-source cooperative observation conditions. The visualization results in Fig.~\ref{fig: weighted_fusion} further confirm the above findings. The longitude and latitude trajectories generated by DAWF are the closest to the ground truth and remain consistent in turning and trajectory-changing regions. Its SOG and COG curves also agree better with the ground truth, showing smoother and more stable variations. The local jumps observed in the COG curves are mainly caused by the angular periodicity effect when the COG crosses the $0^\circ/360^\circ$ boundary. In general, DAWF demonstrates good stability and accuracy in location, SOG, and COG estimation, verifying its effectiveness.

\begin{figure*}[pos=!t]
                \centering
                \includegraphics[width=0.93\textwidth]{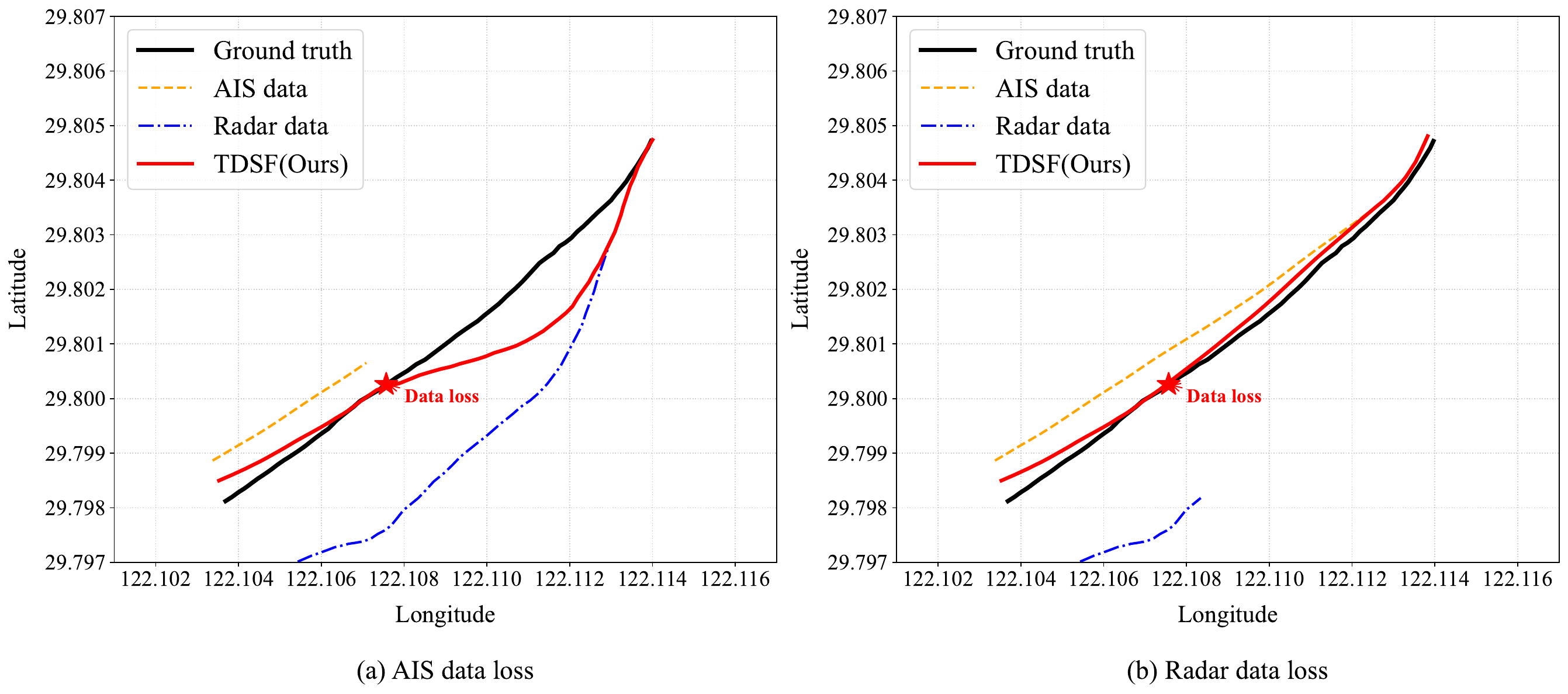}
                \caption{Visualization of different AIS-radar data fusion methods under different simulated scenarios.}
                \label{fig: stitching_fusion}
            \end{figure*}

            To evaluate the robustness of TDSF under dynamic switching between AIS and radar observation modes, we designed two simulated scenarios: Missing AIS and Missing radar. Their performance was compared with that of the direct switching strategy DS, as summarized in Table~\ref{tab: stitching_fusion}. In the Missing AIS scenario, the PJ distance and APD of DS are 210.03 $\mathrm{m}$ and 144.28 $\mathrm{m}$, respectively, whereas TDSF reduces them to 23.20 $\mathrm{m}$ and 37.82 $\mathrm{m}$, corresponding to reductions of 89.0\% and 73.8\%. In the Missing radar scenario, TDSF reduces the PJ distance and APD from 62.04 $\mathrm{m}$ and 33.02 $\mathrm{m}$ to 22.38 $\mathrm{m}$ and 5.75 $\mathrm{m}$, respectively, with an APD reduction of 82.6\%. This improvement is mainly attributed to the use of the fused track before switching as a prior. By combining linear extrapolation with exponential timeliness-decay weighting, TDSF smoothly integrates the historical prediction and the current single-source observation, thereby effectively suppressing abrupt state changes caused by hard switching. The visualization results in Fig.~\ref{fig: stitching_fusion} further show that, compared with the sharp trajectory corners and discontinuities caused by DS at the switching moment, TDSF achieves a smoother trajectory transition while maintaining both trajectory continuity and observation timeliness.

             \begin{table}[pos=!t]
                \centering
                \caption{Performance comparison of different AIS–radar fusion methods on the MAPFusion under simulated scenarios. DS indicates direct switching to single source data. PJ distance indicates position jump distance at the switching moment. APD indicates average position deviation during the missing period.}
                \renewcommand{\arraystretch}{1.35}
                \setlength{\tabcolsep}{3.8pt}
                \label{tab: stitching_fusion}
                \begin{tabular}{ l l c c }
                    \toprule
                    Simulation scenario                 & Method                & PJ distance $\downarrow$              & APD $\downarrow$      \\
                    \midrule
                    \multirow{2}{*}{Missing AIS}        & DS                    & 210.03                                & 144.28                \\
                                                        & TDSF (Ours)           & \textbf{23.20}                        & \textbf{37.82}        \\
                    \midrule
                    \multirow{2}{*}{Missing radar}      & DS                    & 62.04                                 & 33.02                 \\
                                                        & TDSF (Ours)           & \textbf{22.38}                        & \textbf{5.75}         \\
                    \bottomrule
                \end{tabular}
            \end{table}

        \subsubsection{Performance of different object detection methods} 

\begin{figure*}[pos=!t]
                \centering
                \includegraphics[width=0.92\textwidth]{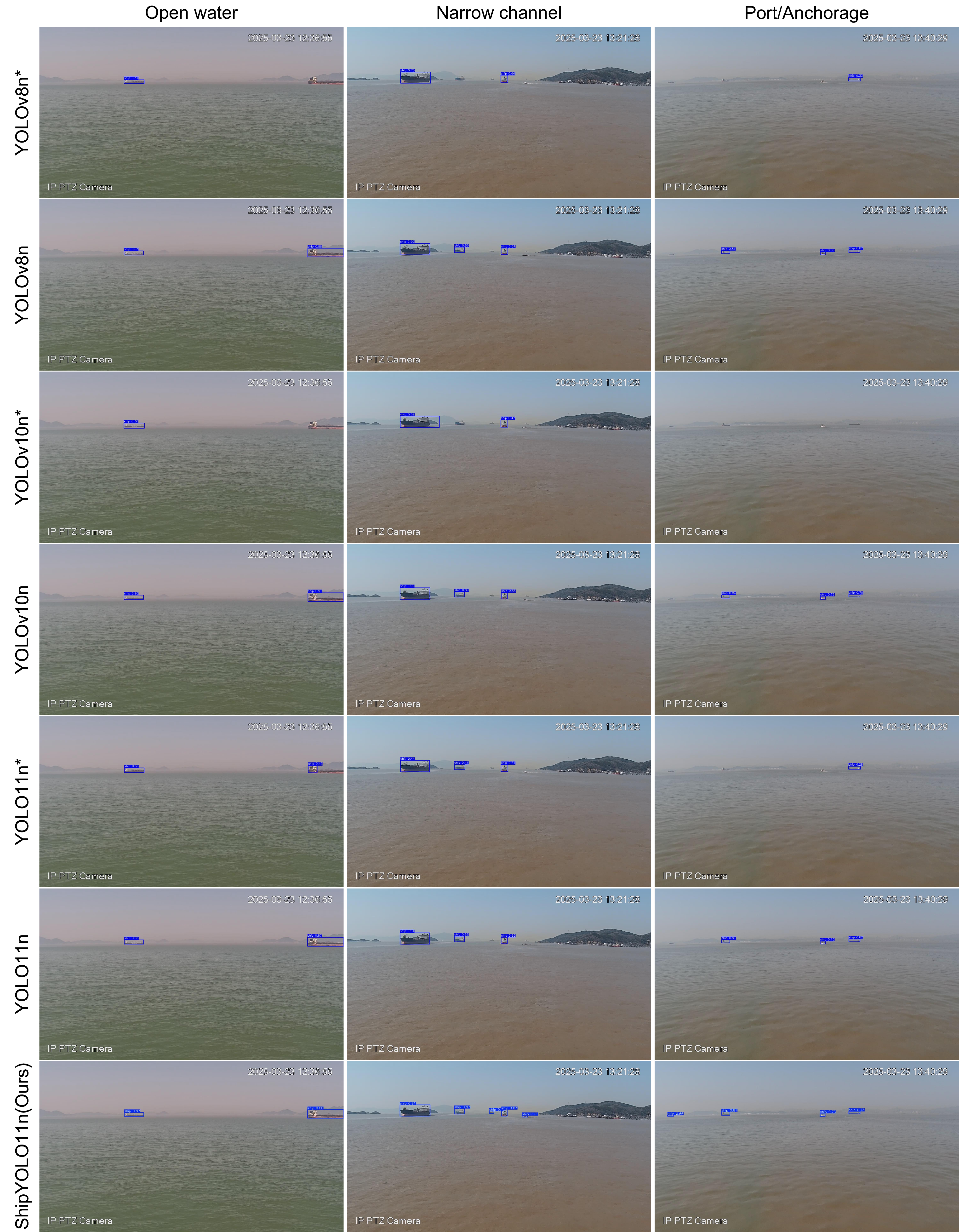}
                \caption{Visualization of results for different object detection methods. * indicates inference using pretrained weights on the COCO dataset. Methods without * indicate inference using weights fine-tuned on the WUTDet dataset.}
                \label{fig: object_detection}
            \end{figure*}

            Table~\ref{tab: object_detection} presents the performance comparison of different object detection methods on MAPFusion. YOLO11n contains only 2.6M parameters and 6.4G FLOPs, and achieves 565.28 FPS with a $640 \times 640$ input, showing good lightweight and real-time inference capability. Considering its overall detection performance using both COCO-pretrained weights and WUTDet fine-tuned weights, we selected YOLO11n as the shipborne visual object detection model. When directly using COCO-pretrained weights, all YOLO models achieve relatively low detection accuracy on MAPFusion, indicating a clear domain gap between general datasets and real maritime ship scenarios. After fine-tuning on WUTDet, the $mAP_{50}$ and $mAP_{50:95}$ of YOLO11n increase from 58.37\% and 28.55\% to 91.03\% and 55.87\%, respectively, demonstrating that scenario-specific fine-tuning can effectively enhance the feature representation capability for ship objects. On this basis, ShipYOLO11n adopts a high-resolution input of $1920 \times 1080$ to preserve detailed information of small and distant ship objects. It achieves the highest $mAP_{50}$ of 94.01\% and $mAP_{50:95}$ of 62.97\%, outperforming the fine-tuned YOLO11n by 2.98\% and 7.10\%, respectively. Although its FPS decreases to 77.19, it is still much higher than the 1 Hz frame-sampling requirement of the NavEYE system. In the narrow channel and port/anchorage scenarios, many distant small objects are present. Fig.~\ref{fig: object_detection} shows that ShipYOLO11n reduces missed and false detections and produces bounding boxes that better fit ship contours. These results indicate that ShipYOLO11n achieves a good balance between detection accuracy and real-time performance.

            \begin{table*}[pos=!t]
                \centering
                \caption{Performance comparison of different object detection methods on the MAPFusion. * indicates inference using pretrained weights on the COCO dataset. Methods without * indicate inference using weights fine-tuned on the WUTDet dataset.}
                \renewcommand{\arraystretch}{1.40}
                \setlength{\tabcolsep}{8pt}
                \label{tab: object_detection}
                \begin{tabular}{ l c c c c c c }
                \toprule
                Method                                                   & Params(M)     & GFLOPs        & Input size        & $mAP_{50}(\%)$ $\uparrow$         & $mAP_{50:95}(\%)$ $\uparrow$      & FPS $\uparrow$             \\
                \midrule
                YOLOv8n* \citep{jocher_Ultralytics_YOLO_2023}            & 3.0           & 8.2           & $640\times640$    & 54.58                             & 27.18                             & 350.17                     \\
                YOLOv8n \citep{jocher_Ultralytics_YOLO_2023}             & 3.0           & 8.2           & $640\times640$    & 91.23                             & 56.16                             & 350.17                     \\
                YOLOv10n* \citep{wang2024yolov10}                        & 2.7           & 8.4           & $640\times640$    & 38.20                             & 20.40                             & 482.84                     \\
                YOLOv10n \citep{wang2024yolov10}                         & 2.7           & 8.4           & $640\times640$    & 89.54                             & 54.24                             & 482.84                     \\
                YOLO11n* \citep{jocher_Ultralytics_YOLO_2023}            & 2.6           & 6.4           & $640\times640$    & 58.37                             & 28.55                             & \textbf{565.28}            \\
                YOLO11n \citep{jocher_Ultralytics_YOLO_2023}             & 2.6           & 6.4           & $640\times640$    & 91.03                             & 55.87                             & \textbf{565.28}            \\
                ShipYOLO11n (Ours)                                       & 2.6           & 32.1          & $1920\times1080$  & \textbf{94.01}                    & \textbf{62.97}                    & 77.19                      \\
                \bottomrule
                \end{tabular}
            \end{table*}

            \begin{table*}[pos=!t]
            \centering
            \caption{Performance comparison of different vision–AIS–radar data association methods on the MAPFusion.}
            \renewcommand{\arraystretch}{1.40}
            \setlength{\tabcolsep}{15pt}
            \label{tab:  vision-ais-radar}
            \begin{tabular}{ l c c c c }
                \toprule
                \multirow{2}{*}{Method}                 & \multicolumn{4}{c}{$Acc\_association (\%)$ $\uparrow$}                                \\
                                                        \cmidrule(lr){2-5}
                                                        & Open water        & Narrow channel        & Port/Anchorage        & All               \\
                \midrule
                Coordinate projection                   & 75.04             & 22.42                 & 16.26                 & 34.49             \\
                earing                                  & 75.57             & 12.56                 & 18.24                 & 32.25             \\
                VARF (Ours)                             & \textbf{96.54}    & \textbf{74.89}        & \textbf{80.37}        & \textbf{83.07}    \\
                \bottomrule
            \end{tabular}
        \end{table*}

        \begin{table*}[pos=!t]
            \centering
            \caption{Ablation study of the MCGA method for AIS-radar data association on the MAPFusion. Baseline denotes the association method using only longitude and latitude constraints, while SCA and CCA represent the association methods that further incorporate SOG and COG constraints based on the Baseline, respectively.}
            \renewcommand{\arraystretch}{1.40}
            \setlength{\tabcolsep}{5pt}
            \label{tab: mcga_ablation}
            \begin{tabular}{ l c c c c c c }
                \toprule
                \multirow{2}{*}{Method}                 & \multirow{2}{*}{SOG}              & \multirow{2}{*}{COG}          & \multicolumn{4}{c}{Precision (\%) $\uparrow$}                                                              \\
                                                                                                                                \cmidrule(lr){4-7}
                                                        &                                   &                                   & Open water        & Narrow channel        & Port/Anchorage        & All                                \\
                \midrule
                Baseline                                & -                                 & -                                 & 70.48             & 58.93                 & 51.35                 & 62.94                              \\
                SCA                                     & +                                 & -                                 & 75.33             & 74.19                 & 65.10                 & 72.68                              \\
                CCA                                     & -                                 & +                                 & 80.59             & 79.40                 & 73.89                 & 78.82                              \\   
                MCGA (Ours)                             & +                                 & +                                 & \textbf{83.24}    & \textbf{84.60}        & \textbf{80.52}        & \textbf{83.26}                     \\
            \bottomrule
            \end{tabular}
        \end{table*}

        \subsubsection{Performance of different vision-AIS–radar data fusion methods} 

\begin{figure*}[pos=!t]
            \centering
            \includegraphics[width=0.92\textwidth]{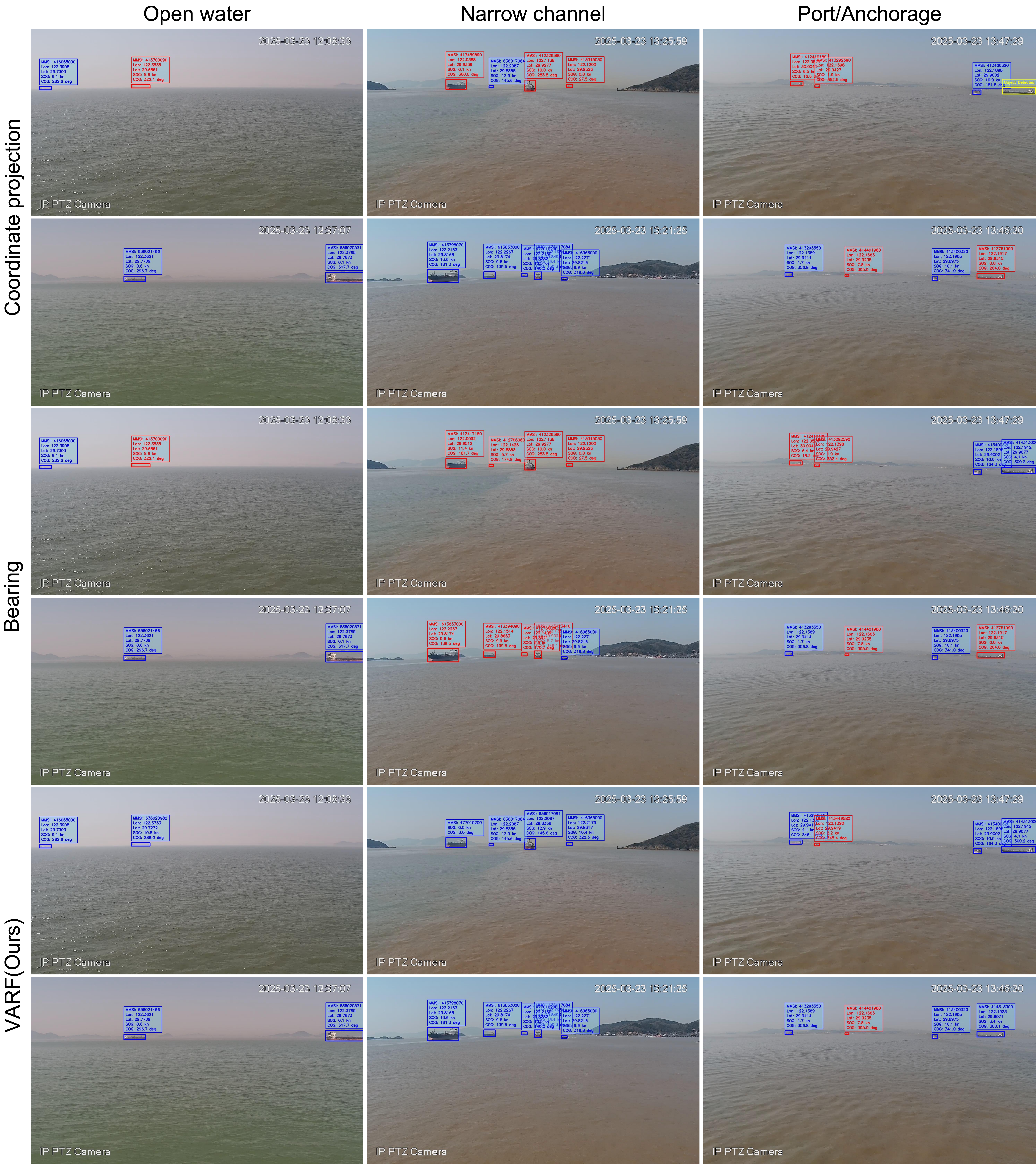}
            \caption{Visualization of results for different vision–AIS–radar data association methods. Blue indicates correct associations, red indicates incorrect associations, and yellow indicates detected objects that are not successfully associated.}
            \label{fig: vision-ais-radar_fusion}
        \end{figure*}

        To validate the effectiveness of VARF, Table~\ref{tab:  vision-ais-radar} compares it with Coordinate projection and the bearing method. VARF achieves the highest association accuracy of 83.07\% across all scenarios, outperforming coordinate projection (34.49\%) and the bearing method (32.25\%) by 48.58\% and 50.82\%, respectively. This indicates that the joint bearing-distance constraint can effectively improve the reliability of vision-AIS-radar cross-modal association. The performance differences across scenarios are mainly affected by object density, spatial distribution, and viewpoint variation. In the open water scenario, objects are fewer and more widely separated. Therefore, coordinate projection and the bearing-based method can still achieve an accuracy of about 75\%. However, due to camera installation errors and the limitation of using only a single bearing constraint, their performance remains lower than that of VARF, which reaches 96.54\%. In the narrow channel and port/anchorage scenarios, ship crossings and near-collinear object distributions occur more frequently, making single projection or bearing constraints prone to matching ambiguity. In contrast, VARF jointly uses normalized bearing and distance deviations to effectively alleviate association ambiguity among objects with similar bearings, and therefore maintains accuracies of 74.89\% and 80.37\%, respectively. As shown in Fig.~\ref{fig: vision-ais-radar_fusion}, all three methods achieve relatively good association results in the open water scenario, although Coordinate projection and the bearing-based method still produce a few incorrect associations. In the narrow channel and port/anchorage scenarios, due to dense ship objects and the presence of objects with similar bearings, the first two methods are more likely to produce incorrect or missing associations. In comparison, VARF obtains more correct association results across different scenarios, demonstrating stronger robustness and practical applicability.

    \subsection{Ablation experiments}

        Table~\ref{tab: mcga_ablation} presents the ablation results of the MCGA method. The baseline using only longitude and latitude constraints achieves a precision of 62.94\%. After introducing the SOG constraint, the precision of SCA increases to 72.68\% (9.74\%). This indicates that SOG information helps distinguish objects that are spatially close but have different motion states. After further introducing the COG constraint, CCA achieves a Precision of 78.82\% (15.88\%). Compared with SOG, COG can more directly reflect the motion trend of an object, and therefore provides better discrimination for crossing, parallel, and closely spaced objects. Across different scenarios, the performance differences among methods are relatively small in the open water scenario. However, in the narrow channel and port/anchorage scenarios, as object density and crossing situations increase, the precision of the baseline decreases to 58.93\% and 51.35\%, respectively. In contrast, MCGA jointly uses position, SOG, and COG constraints, achieving the best precision values of 83.24\%, 84.60\%, and 80.52\% in the three scenarios, respectively. In addition, the precision fluctuation of MCGA across different scenarios is only 4.08\%, which is significantly lower than the 19.13\% of the baseline.

        \begin{table*}[pos=!t]
            \centering
            \caption{Ablation study of the VARF method on the MAPFusion.}
            \renewcommand{\arraystretch}{1.40}
            \setlength{\tabcolsep}{5pt}
            \label{tab: varf_ablation}
            \begin{tabular}{ l l c c c c }
                \toprule
                \multirow{2}{*}{Method}         & \multirow{2}{*}{Assignment strategy}      & \multicolumn{4}{c}{Acc\_association (\%) $\uparrow$}                                              \\
                                                                                            \cmidrule(lr){3-6}
                                                &                                           & Open water            & Narrow channel            & Port/Anchorage            & All               \\
                \midrule
                bearing + NN                    & Nearest neighbor                          & 75.57                 & 14.01                     & 15.09                     & 31.46             \\
                bearing + Hungarian             & Hungarian                                 & 75.57                 & 12.56                     & 18.24                     & 32.25             \\
                distance + bearing + NN         & Nearest Neighbor                          & 96.54                 & 68.55                     & 60.69                     & 73.13             \\
                VARF (Ours)                     & Hungarian                                 & \textbf{96.54}        & \textbf{74.89}            & \textbf{80.37}            & \textbf{83.07}    \\
                \bottomrule
            \end{tabular}
        \end{table*}

         \begin{figure*}[pos=!t]
            \centering
            \includegraphics[width=0.92\textwidth]{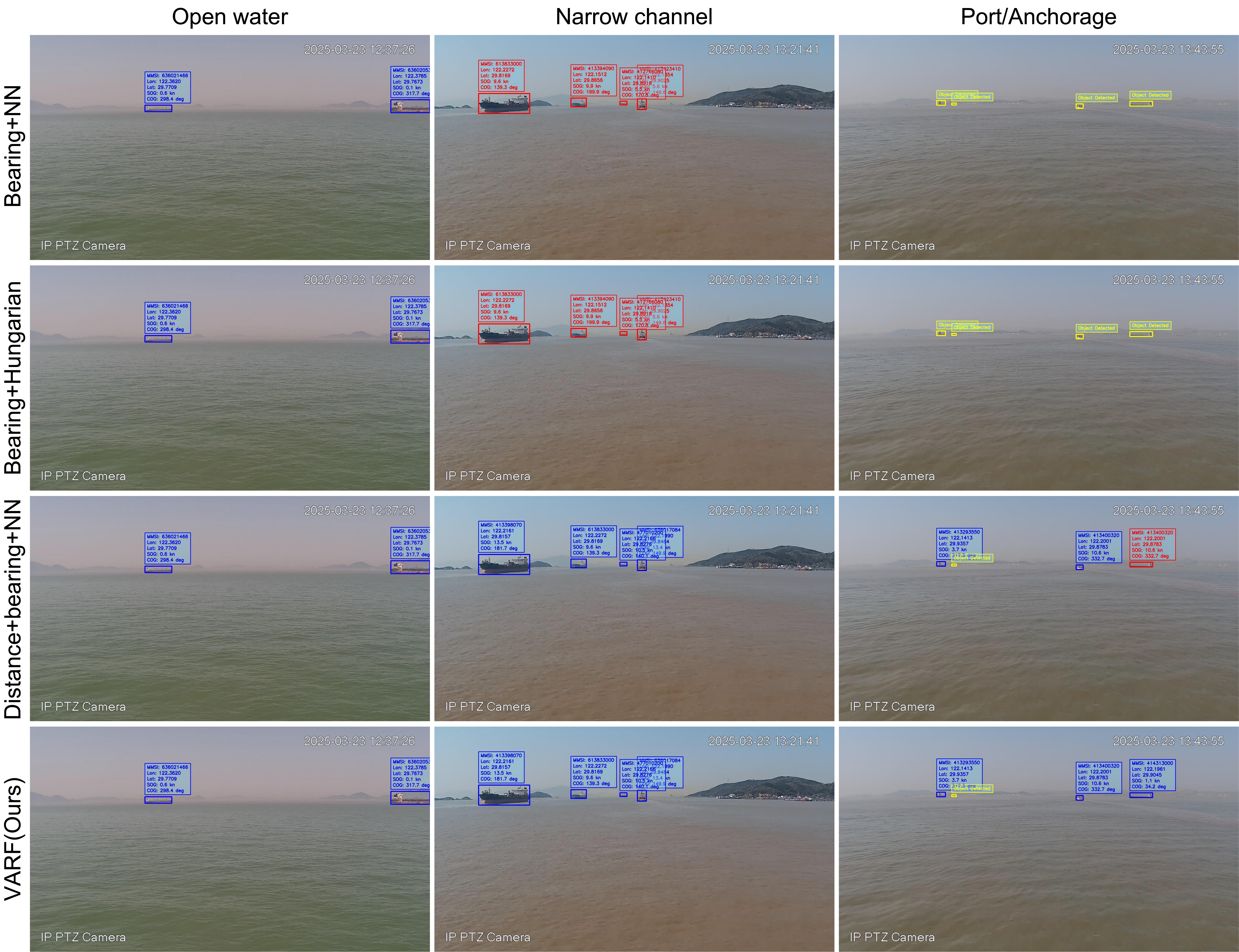}
            \caption{Visualization of ablation results for vision-AIS-radar data association under different feature constraints and matching strategies. Blue indicates correct associations, red indicates incorrect associations, and yellow indicates detected objects that are not successfully associated.}
            \label{fig:  vision-ais-radar_fusion_ablation}
        \end{figure*}

        Table~\ref{tab: varf_ablation} presents the ablation results of the VARF method. When only the bearing feature is used, bearing + Hungarian brings only a limited improvement over bearing + NN. This indicates that the effect of global optimal assignment is limited when the feature discriminability is insufficient. After introducing the distance feature, the overall $Acc_{association}$ is significantly improved, showing that distance information can effectively alleviate association ambiguity caused by objects with similar bearings. This effect is particularly evident in high-density scenarios such as narrow channel and port/anchorage. In the open water scenario, range + bearing + NN and VARF both achieve an accuracy of 96.54\%, indicating that the marginal contribution of the assignment strategy is limited in low-density scenarios. In more complex scenarios, however, the combination of discriminative features and global optimal assignment can further improve association performance. The visualization results in Fig.~\ref{fig:  vision-ais-radar_fusion_ablation} also show that methods relying only on bearing are prone to incorrect associations in Narrow channel and unassociated objects in Port/Anchorage. In contrast, VARF achieves more stable and accurate association results across all three scenarios.

\section{Our NavEYE system} \label{sec: NavEYE}

    Based on the methods proposed in Section~\ref{sec: method}, we developed a vision-centered multi-sensor fusion-based situational awareness system for ISVs, named NavEYE. The system enables end-to-end processing and visual representation from multi-source sensor observations to navigation assistance information. NavEYE consists of two main parts: hardware and software.

    \subsection{Hardware}

\begin{figure*}[pos=!t]
            \centering
            \includegraphics[width=0.92\textwidth]{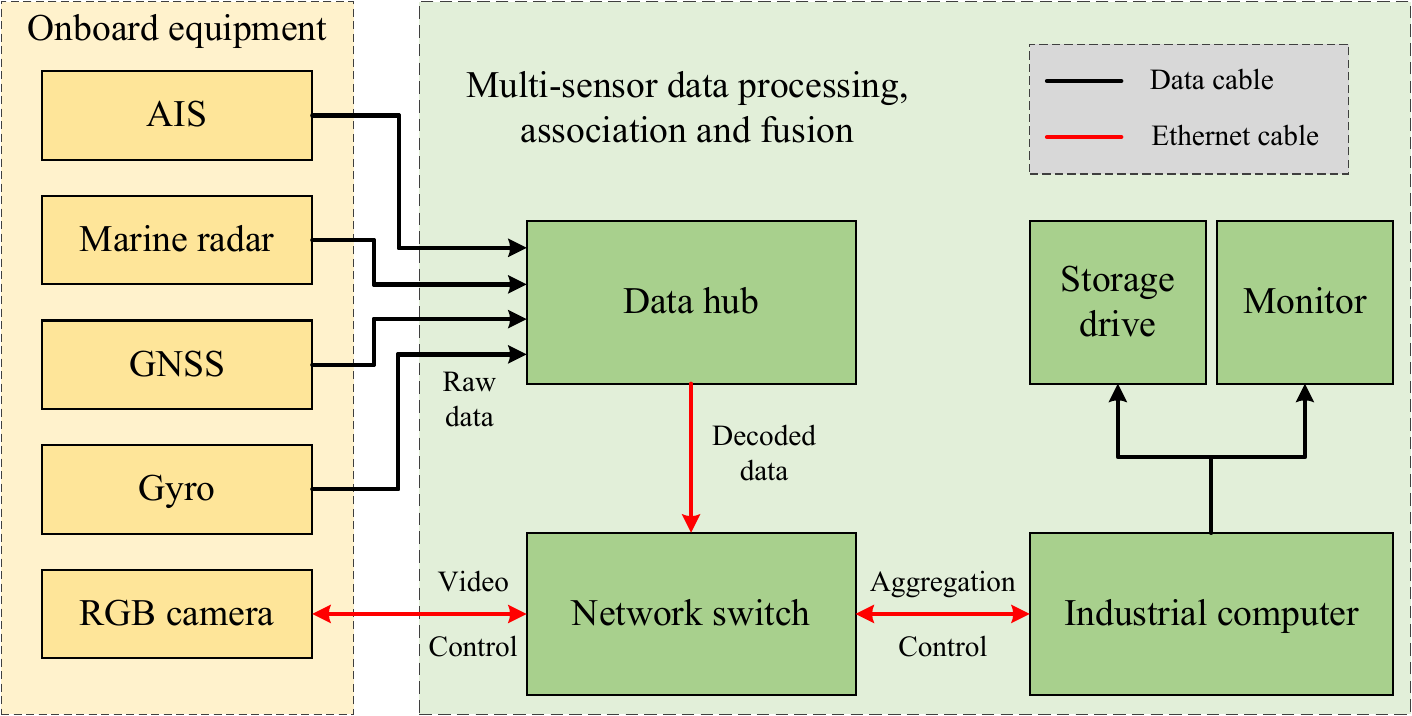}
            \caption{The hardware architecture of NavEYE. AIS, radar, GNSS, and gyrocompass data are collected by onboard equipment, decoded by the data hub, and transmitted to the industrial computer through the network switch, while RGB camera video data are directly connected to the industrial computer via the network switch. Subsequently, the industrial computer performs multi-source data fusion.}
            \label{fig: system_hardware_connection}
        \end{figure*}

        Fig.~\ref{fig: system_hardware_connection} shows the hardware architecture of NavEYE. The system consists of marine radar, AIS, GNSS, gyrocompass, RGB camera, data hub, network switch, industrial computer, monitor, and storage devices. These hardware units are interconnected through data cables or ethernet cables to form a local area network, enabling synchronized acquisition and transmission of multi-source sensing data. Among them, the data hub and the industrial computer serve as the core hardware platforms of the system. The data hub runs the Ubuntu operating system and is equipped with a Cortex-A9 quad-core 1.4~GHz CPU, 1~GB RAM, and 8~GB eMMC storage. It is mainly responsible for decoding and preprocessing data from marine radar, AIS, GNSS, and the gyrocompass (Fig.~\ref{fig: method_overview}). The industrial computer runs the Windows operating system and is equipped with an Intel\textsuperscript{\textregistered} Core\textsuperscript{TM} i5-8400 CPU at 2.80~GHz, 16~GB RAM, and an NVIDIA GeForce RTX 4060 GPU with 8~GB memory. It is used to perform core tasks such as object detection, data association, data fusion, and deployment of the NavEYE system. The storage device is used to record multi-source perception data collected during system testing. The gyrocompass provides real-time heading information of the own ship, while GNSS provides high-precision longitude, latitude, SOG, and COG information, offering basic navigation information for multi-sensor spatial alignment and object association.

        The workflow of the system is as follows. First, the raw messages output by marine radar, AIS, GNSS, and the gyrocompass are transmitted to the data hub. The data hub parses and preprocesses these messages to obtain structured perception data, where the marine radar and AIS data are organized as lists, while the GNSS and gyrocompass data are organized as dictionaries. The processed data are then pushed through the message queuing telemetry transport (MQTT) protocol. Subsequently, these data, together with the synchronized video stream, are transmitted to the industrial computer through the network switch. The NavEYE system then performs multi-source information inference and fusion computation. Finally, the processed results are sent to the monitor for real-time visual monitoring. To support subsequent method optimization and system performance evaluation, the video streams collected during NavEYE testing and the sensor data processed by the data hub are synchronously stored in the storage device in real time. In addition, the industrial computer supports adjustment of the pitch angle and horizontal alignment of the RGB camera.

    \subsection{Software}

        To alleviate the computational burden of frame-by-frame video processing, a video preprocessing mechanism was introduced at the data input stage to enhance overall system efficiency. Specifically, raw video streams were first sampled at 1 Hz; the sampled frames were subsequently encoded into binary image data and transmitted to the server for further processing via the gRPC protocol. This preprocessing pipeline was implemented through a Java-based gRPC client, ensuring efficient and reliable data transmission. Moreover, gRPC supports cross-language client–server communication, which improves system scalability and interoperability.

        \begin{figure*}[pos=!t]
            \centering
            \includegraphics[width=0.92\textwidth]{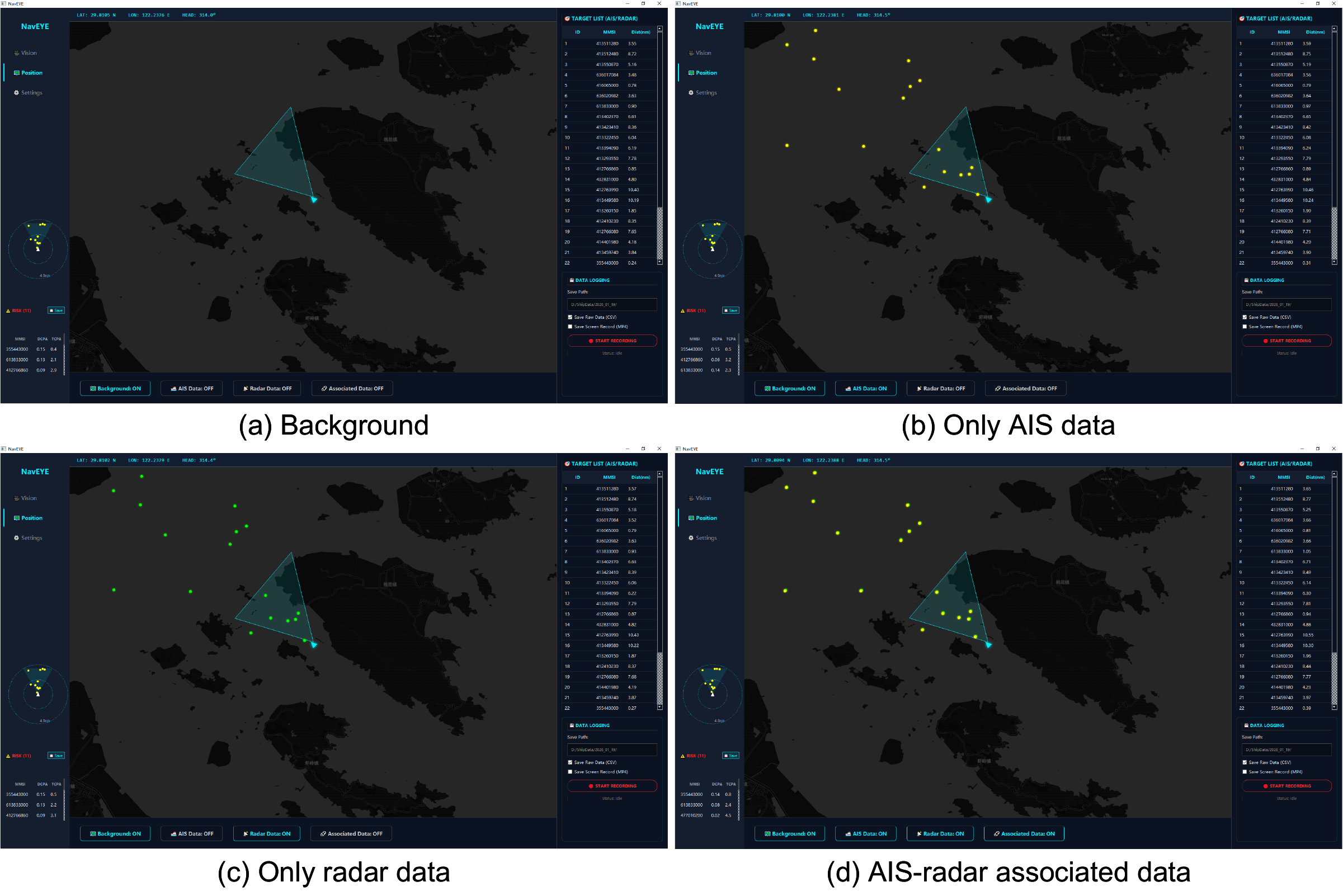}
            \caption{Fused perception and anomaly monitoring in the NavEYE vision panel.Fusion perception and anomaly monitoring functions in the NavEYE vision panel. From left to right: (a) visualization of vision–AIS–radar fusion results, including information such as MMSI, longitude, latitude, SOG, and COG; (b) identification and warning display of vessels with collision risk.}
            \label{fig: vision_panel}
        \end{figure*}

        \begin{figure*}[pos=!t]
            \centering
            \includegraphics[width=0.92\textwidth]{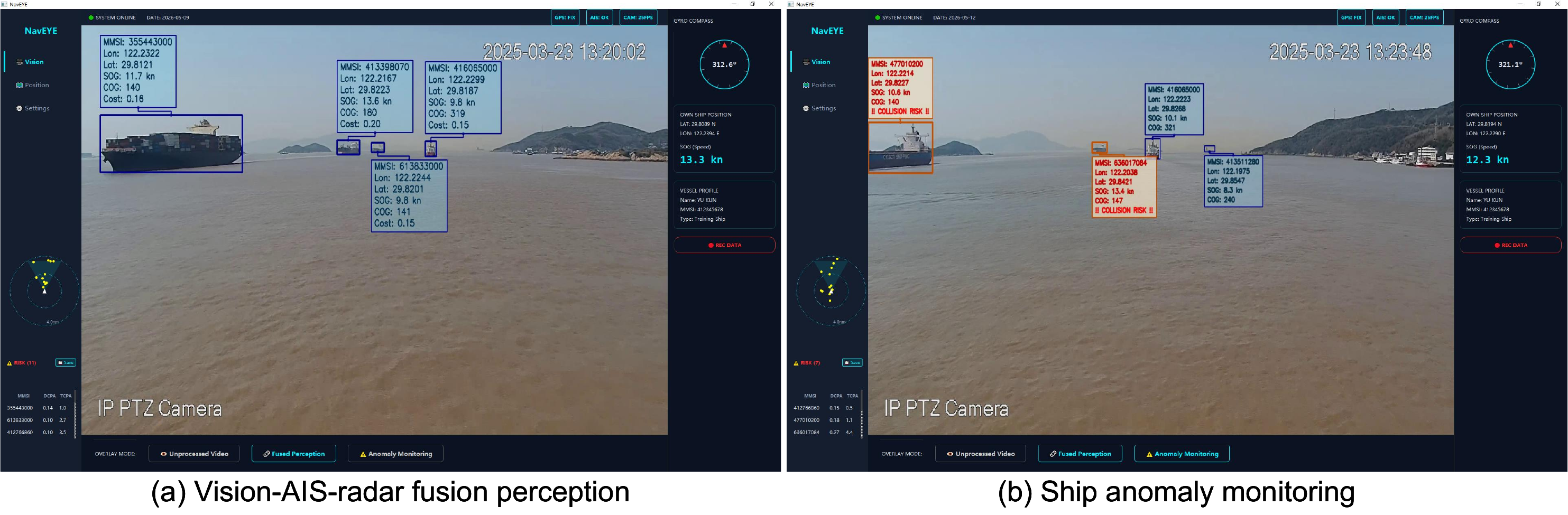}
            \caption{NavEYE position panel with layer control functionality. From top left to bottom right: (a) display of the own ship position, COG, and background map; (b) AIS data overlaid on (a); (c) radar data overlaid on (a); and (d) AIS–radar associated data overlaid on (a).}
            \label{fig: position_panel}
        \end{figure*}

        The video preprocessing produces binary image data. The other sensor data are preprocessed into radar and AIS data in list form, as well as GNSS and gyrocompass data in dictionary form. All these data are then input into the vision-centered vision-AIS-radar multi-sensor data association and fusion framework (Section~\ref{sec: method}). To efficiently process the multi-source heterogeneous data, we implement the framework using a multiprocessing mechanism. NavEYE consists of three main processes. The first is the MQTT data reception process, which receives the multi-source sensor data pushed by the data hub at one-second intervals. The second is the gRPC server process, which is responsible for the core computational workflow, including the reception and decoding of binary images, AIS-radar data preprocessing (Fig.~\ref{fig: method_overview}), AIS-radar data association (Section~\ref{sec: aIS-radar data association}), AIS-radar data fusion (Section~\ref{sec: adaptive AIS-radar data fusion}), object detection model inference (Section~\ref{sec: vision–AIS–radar data fusion based on bearing and distance}), and vision-AIS-radar data association (Section~\ref{sec: vision–AIS–radar data fusion based on bearing and distance}). The processing results are then returned to the client through the gRPC protocol. The third is the system management process, which schedules the MQTT reception process and the gRPC server process and monitors their running status. The framework is implemented in Python, and the core model inference module is based on PyTorch.

    \subsection{Interface}

        To validate the engineering feasibility and practical applicability of the NavEYE system, we evaluate it using the MAPFusion. The system simultaneously integrates multi-source sensor data, including AIS, radar, GNSS, gyrocompass, and RGB camera inputs, and performs end-to-end processing and visualization from raw perception data to navigation assistance information on an industrial computer platform. Using GNSS and gyrocompass measurements, the system continuously acquires the own ship’s longitude, latitude, SOG, and COG in real time. These navigation states are dynamically displayed through digital dashboards and simulated compass graphical components, enabling operators to maintain accurate awareness of the ship’s current navigation status. On this basis, the system consists of two core interfaces: the Vision Panel and the Position Panel. The Vision Panel provides video-stream-based fusion perception and abnormal event warning visualization, while the Position Panel enables multi-layer object monitoring based on a two-dimensional electronic chart.

        Fig.~\ref{fig: vision_panel} illustrates the fusion perception and abnormal warning visualization results of the NavEYE system Vision Panel during operation. The system performs multi-source data association and fusion for ship objects detected in video streams, and overlays structured labels above each object in real time, including MMSI, longitude and latitude, SOG, COG, and collision risk coefficient. Meanwhile, the NavEYE system automatically determines collision risk levels based on the calculated DCPA and TCPA values. Objects within safe conditions are rendered with blue bounding boxes, whereas objects with potential collision risks are immediately highlighted using orange warning indicators. The information of high-risk objects is simultaneously synchronized to the miniature radar view and the risk object list on the left side of the interface, providing operators with intuitive multi-dimensional situational awareness assistance.

        Fig.~\ref{fig: position_panel} presents the multi-layer interactive visualization of the NavEYE system Position Panel. Centered on the real-time GNSS position of the own ship, the NavEYE system renders a dark-themed electronic chart using the Leaflet open-source framework and supports independent layer switching for AIS data, radar data, and AIS–radar fusion data. AIS objects are represented by yellow circular markers, while radar objects are represented by green circular markers. When the fusion layer is enabled, successfully associated AIS and radar objects are spatially overlaid and rendered together. Meanwhile, the object list on the right side of the interface is synchronously updated with navigation-related information, including object ID, MMSI, and relative distance. The own ship is represented by a cyan triangular icon together with a sector-shaped region indicating the real-time camera field of view, providing an intuitive correspondence between visual space and geographic space.
        
\section{Conclusion and discussion} \label{sec: conclusions and discussion}
    
    To address the problem of multi-source sensor data fusion in ISV situational awareness, we propose a vision-centered vision-AIS-radar multi-sensor fusion framework and develop NavEYE based on this framework. For AIS–radar data association, the proposed MCGA method integrates position, SOG, and COG features to construct an association cost matrix, and employs Mahalanobis distance gating together with the Hungarian algorithm to achieve globally optimal association. This improves association accuracy and robustness in dense object scenarios. For AIS–radar data fusion, the proposed DAWF and TDSF methods address the distance-dependent characteristics of radar measurement errors and the dual-/single-source observation switching problem, respectively. They enable adaptive fusion weight allocation and smooth transition between tracking modes, thereby improving trajectory estimation accuracy and continuity. For vision-AIS-radar data fusion, the proposed VARF method models cross-modal similarity using both bearing and distance constraints, and applies the Hungarian algorithm to obtain globally optimal association. This effectively alleviates association ambiguity in dense multi-object scenarios. Experimental results on the constructed MAPFusion dataset demonstrate that the proposed methods outperform existing approaches in AIS–radar association, AIS–radar fusion, and vision–AIS–radar fusion tasks. In addition, the developed NavEYE system enables end-to-end processing and visual presentation from raw observations to navigation-assistance information.

    However, MAPFusion and the VARF method still have certain limitations. MAPFusion is mainly designed for normal visibility and illumination conditions, and lacks complex environmental data such as low visibility, rain, fog, and strong backlight. In VARF, the bottom position of the visual bounding box is used as a visual distance feature. This setting may be affected by camera pitch variations, wave-induced occlusion, and object-scale differences, thereby degrading the association performance. In future work, we will further expand the scenario coverage of MAPFusion and explore the integration of IMU-based attitude compensation and visual distance modeling to improve the robustness of cross-modal association.





\printcredits

\section*{Declaration of competing interest}
The authors declare that they have no known competing financial interests or personal relationships that could have appeared to influence the work reported in this paper.

\section*{Acknowledgments}
This work was supported by the National Natural Science Foundation of China (Nos. 52422111 and 52271365), the Excellent Youth Foundation of Hubei Scientific Committee (No. 2024AFA042), and the Key Research and Development Program of Hainan Province (No. ZDYF2026GXJS024).

\bibliographystyle{cas-model2-names}

\bibliography{main}

@article{liang2026wutdet,
  title   = {WUTDet: A 100K-Scale Ship Detection Dataset and Benchmarks with Dense Small Objects},
  author  = {Liang, Junxiong and Bao, Mengwei and Wang, Tianxiang and Wang, Xinggang and Liu, An-An and Liu, Ryan Wen},
  journal = {arXiv preprint arXiv:2604.07759},
  year    = {2026}
}

@article{fortmann1983sonar,
  title   = {Sonar tracking of multiple targets using joint probabilistic data association},
  author  = {Fortmann, Thomas E. and Bar-Shalom, Yaakov and Scheffe, Molly},
  journal = {IEEE J. Oceanic Eng.},
  volume  = {8},
  number  = {3},
  pages   = {173-184},
  year    = {1983},
}

@article{blom1988interacting,
  title   = {The interacting multiple model algorithm for systems with Markovian switching coefficients},
  author  = {Blom, Henk A. P. and Bar-Shalom, Yaakov},
  journal = {IEEE Trans. Autom. Control},
  volume  = {33},
  number  = {8},
  pages   = {780-783},
  year    = {1988},
}

@article{reid2003algorithm,
  title={An algorithm for tracking multiple targets},
  author={Reid, Donald},
  journal={IEEE Trans. Autom. Control},
  volume={24},
  number={6},
  pages={843-854},
  year={2003},
  publisher={IEEE}
}

@article{yan2023association,
  title={Association and Fusion of Ship AIS and Radar Track Data},
  author={Yan, Q and Zheng, S and Guan, H and Li, L},
  journal={J. Wuhan Univ. Technol},
  volume={47},
  pages={185-190},
  year={2023}
}

@article{yang2022multitarget,
  title   = {Multi-target association algorithm of AIS-radar tracks using graph matching-based deep neural network},
  author  = {Yang, Yang and Yang, Fan and Sun, Li and Xiang, Tao and Lv, Peng},
  journal = {Ocean Eng.},
  volume  = {266},
  pages   = {112208},
  year    = {2022},
}

@article{wang2023intelligent,
  title   = {Intelligent marine area supervision based on AIS and radar fusion},
  author  = {Wang, Chi Ming and Li, Yanan and Min, Lanxi and Chen, Jiahui and Lin, Zhen and Su, Shuo and Zhang, Yuhan and Chen, Qian and Chen, Yihong and Duan, Xiaoyan and others},
  journal = {Ocean Eng.},
  volume  = {285},
  pages   = {115373},
  year    = {2023},
}

@article{kazimierski2015radar,
  title   = {Radar and Automatic Identification System track fusion in an Electronic Chart Display and Information System},
  author  = {Kazimierski, Witold and Stateczny, Andrzej},
  journal = {J. Navig.},
  volume  = {68},
  number  = {6},
  pages   = {1141-1154},
  year    = {2015},
}

@inproceedings{liu2023bevfusion,
  title     = {BEVFusion: Multi-Task Multi-Sensor Fusion with Unified Bird's-Eye View Representation},
  author    = {Liu, Zhijian and Tang, Haotian and Amini, Alexander and Yang, Xinyu and Mao, Huizi and Rus, Daniela and Han, Song},
  booktitle = {Proc. ICRA},
  pages     = {2774-2781},
  year      = {2023},
  publisher = {IEEE},
}

@inproceedings{nabati2021centerfusion,
  title     = {CenterFusion: Center-Based Radar and Camera Fusion for 3D Object Detection},
  author    = {Nabati, Ramin and Qi, Hairong},
  booktitle = {Proc. CVPR},
  pages     = {1527-1536},
  year      = {2021},
}

@inproceedings{lu2021fusion,
  title     = {Fusion of Camera-Based Vessel Detection and AIS for Maritime Surveillance},
  author    = {Lu, Yongqiang and Ma, Hongjie and Smart, Edward and Vuksanovic, Branislav and Chiverton, John and Radhakrishna Prabhu, Shanker G. and Glaister, Malcolm and Dunston, Eric and Hancock, Chris},
  booktitle = {Proc. ICAC},
  pages     = {1-6},
  year      = {2021},
  publisher = {IEEE},
}

@inproceedings{gulsoylu2024image,
  title     = {Image and AIS Data Fusion Technique for Maritime Computer Vision Applications},
  author    = {G{\"u}lsoylu, Emre and Koch, Paul and Yildiz, Mert and Constapel, Manfred and Kelm, Andr{\'e} Peter},
  booktitle = {Proc. WACV},
  pages     = {859-868},
  year      = {2024}
}

@article{huang2025surface,
  title   = {Surface Vessels Detection and Tracking Method and Datasets with Multi-Source Data Fusion in Real-World Complex Scenarios},
  author  = {Huang, Wei and Feng, Hao and Xu, Haoran and others},
  journal = {Sensors},
  volume  = {25},
  number  = {7},
  pages   = {2179},
  year    = {2025},
}

@article{han2020autonomous,
  title   = {Autonomous collision detection and avoidance for ARAGON USV: Development and field tests},
  author  = {Han, Jiwon and Cho, Younghun and Kim, Jinwhan and Kim, Jeonghyun and Son, Nam-Sun and Kim, Seung-Yong},
  journal = {J. Field Rob.},
  volume  = {37},
  number  = {6},
  pages   = {987-1002},
  year    = {2020},
}

@article{prasad2017maritime,
  title   = {Maritime situational awareness using adaptive multi-sensor management under hazy conditions},
  author  = {Prasad, Dilip K. and Prasath, C. Krishna and Rajan, Deepu and Rachmawati, Lily and Rajabally, Eshan and Quek, Chai},
  journal = {arXiv preprint arXiv:1702.00754},
  year    = {2017}
}

@article{fan2018research,
  title   = {Research on shipborne aided navigation system based on enhanced traffic environment perception},
  author  = {Fan, Yaotian and Huang, Liwen and Jiang, Dan and Xu, Xianzhang},
  journal = {PLOS ONE},
  volume  = {13},
  number  = {10},
  pages   = {e0206402},
  year    = {2018},
}

@article{jovanovic2024review,
  title   = {Review of research progress of autonomous and unmanned shipping},
  author  = {Jovanovi{\'c}, Ivan and others},
  journal = {J. Mar. Eng. Technol.},
  year    = {2024},
}

@article{thombre2022sensors,
  title   = {Sensors and AI Techniques for Situational Awareness in Autonomous Ships: A Review},
  author  = {Thombre, Sarang and Zhao, Zheng and Ramm-Schmidt, Henrik and Vallet Garc{\'i}a, Jos{\'e} M. and Malkam{\"a}ki, Tuomo and Nikolskiy, Sergey and Hammarberg, Toni and Nuortie, Hiski and Bhuiyan, M. Zahidul H. and S{\"a}rkk{\"a}, Simo and Lehtola, Ville V.},
  journal = {IEEE Trans. Intell. Transp. Syst.},
  volume  = {23},
  number  = {1},
  pages   = {64-83},
  year    = {2022},
}

@article{lee2023data,
  title   = {Data association for autonomous ships based on virtual simulation environment},
  author  = {Lee, Hye-Won and Roh, Myung-Il and Cho, Yeong-Min and Park, Jeong-Ho},
  journal = {Ocean Eng.},
  volume  = {281},
  pages   = {114646},
  year    = {2023},
}

@article{harati2007ais,
  title   = {Automatic Identification System AIS: Data Reliability and Human Error Implications},
  author  = {Harati-Mokhtari, Abbas and Wall, Alan and Brooks, Philip and Wang, Jin},
  journal = {J. Navig.},
  volume  = {60},
  number  = {3},
  pages   = {373-389},
  year    = {2007},
}

@article{bounaceur2022analysis,
  title   = {Analysis of Small Sea-Surface Targets Detection Performance According to Airborne Radar Parameters in Abnormal Weather Environments},
  author  = {Bounaceur, Halim and Bourlier, Christophe and Sergievskaya, Irina A. and Fabbro, Val{\'e}rie},
  journal = {Sensors},
  volume  = {22},
  number  = {9},
  pages   = {3263},
  year    = {2022},
}

@article{guo2023asynchronous,
  title   = {Asynchronous Trajectory Matching-Based Multimodal Maritime Data Fusion for Vessel Traffic Surveillance in Inland Waterways},
  author  = {Guo, Yuxin and Liu, Ryan Wen and Qu, Jingxuan and Lu, Yongqiang and Zhu, Fei and Lv, Yisheng},
  journal = {IEEE Trans. Intell. Transp. Syst.},
  volume  = {24},
  number  = {11},
  pages   = {12779-12792},
  year    = {2023},
}

@article{qiao2021marine,
  title   = {Marine Vision-Based Situational Awareness Using Discriminative Deep Learning: A Survey},
  author  = {Qiao, Dong and Liu, Guoyuan and Zhang, Jialun and Yang, Yamin},
  journal = {J. Mar. Sci. Eng.},
  volume  = {9},
  number  = {4},
  pages   = {397},
  year    = {2021},
}

@article{sinnott1984virtues,
  title={Virtues of the Haversine},
  author={Sinnott, Roger W},
  journal={Sky and telescope},
  volume={68},
  number={2},
  pages={158},
  year={1984}
}

@article{ju1982control,
  title={Control problems of grey systems},
  author={Ju-Long, Deng},
  journal={Syst. Control Lett.},
  volume={1},
  number={5},
  pages={288-294},
  year={1982},
  publisher={Elsevier}
}

@article{wang2024yolov10,
  title={Yolov10: Real-time end-to-end object detection},
  author={Wang, Ao and Chen, Hui and Liu, Lihao and Chen, Kai and Lin, Zijia and Han, Jungong and Ding, Guiguang},
  journal={Advances in neural information processing systems},
  volume={37},
  pages={107984-108011},
  year={2024}
}

@misc{jocher_Ultralytics_YOLO_2023,
  author = {Jocher, Glenn and Qiu, Jing and Chaurasia, Ayush},
  license = {AGPL-3.0},
  month = jan,
  title = {Ultralytics YOLO},
  url = {https://github.com/ultralytics/ultralytics},
  version = {8.0.0},
  year = {2023}
}

@article{zhang2025multi,
  title={Multi-Sensor Data Fusion Meets Edge Computing for Intelligent Surface Vehicles},
  author={Zhang, Boxing and Liu, Ryan Wen and Liu, Jingxian and Chui, Kwok Tai and Gupta, Brij B},
  journal={IEEE Internet Things Mag.},
  volume={8},
  number={5},
  pages={127-135},
  year={2025},
  publisher={IEEE}
}

@article{lu2026graph,
  title={Graph learning-driven multi-vessel association: fusing multimodal data for maritime intelligence},
  author={Lu, Yuxu and Yang, Kaisen and Yang, Dong and Ding, Haifeng and Weng, Jinxian and Liu, Ryan Wen},
  journal={IEEE Trans. Intell. Transp. Syst.},
  year={2026},
  publisher={IEEE}
}

@article{liu2022intelligent,
  title={Intelligent edge-enabled efficient multi-source data fusion for autonomous surface vehicles in maritime internet of things},
  author={Liu, Ryan Wen and Guo, Yu and Nie, Jiangtian and Hu, Qin and Xiong, Zehui and Yu, Han and Guizani, Mohsen},
  journal={IEEE Trans. Green Commun. Networking},
  volume={6},
  number={3},
  pages={1574-1587},
  year={2022},
  publisher={IEEE}
}

@article{yang2025tad,
  title={TAD-YOLO: A lightweight nearshore ship detection method for small USVs in maritime trade surveillance},
  author={Yang, Xiaofei and Shao, Shuo and Liu, Weiyong and Xiang, Zhengrong and Zhang, Bin},
  journal={Ocean Eng.},
  volume={341},
  pages={122618},
  year={2025},
  publisher={Elsevier}
}

@article{fang2026orientation,
  title={Orientation-aware dynamic vision-AIS association for UAV-based maritime surveillance},
  author={Fang, Ao and Yuan, Haiwen and Zheng, Kuanlei and Zhang, Bulin and Xiao, Changshi and Li, Qiliang},
  journal={Ocean Eng.},
  volume={358},
  pages={125796},
  year={2026},
  publisher={Elsevier}
}

@article{xu2026challenges,
  title={Challenges for the Development of Maritime Autonomous Surface Ships},
  author={Xu, Haitong and Soares, C Guedes},
  journal={Autonomous Transportation Research},
  volume={2},
  pages={1-24},
  year={2026},
  publisher={Elsevier}
}

@article{liu2025concepts,
  title={Concepts, Key Technologies, Applications and Development Trends in Autonomous Transportation Systems},
  author={Liu, Jiongjiong and Yu, Haiyang and Huang, Ailing and Ma, Xiaoping and Wu, Bing and Sun, Jian and Jia, Limin and Chen, Yong and Wang, Yunpeng and Wang, Jin and others},
  journal={Autonomous Transportation Research},
  volume={1},
  pages={1-23},
  year={2025},
  publisher={Elsevier}
}

@article{liu2023aioenet,
  title={AiOENet: All-in-one low-visibility enhancement to improve visual perception for intelligent marine vehicles under severe weather conditions},
  author={Liu, Ryan Wen and Lu, Yuxu and Guo, Yu and Ren, Wenqi and Zhu, Fenghua and Lv, Yisheng},
  journal={IEEE Trans. Intell. Veh.},
  volume={9},
  number={2},
  pages={3811--3826},
  year={2023},
  publisher={IEEE}
}

@article{liu2024real,
  title={Real-time multi-scene visibility enhancement for promoting navigational safety of vessels under complex weather conditions},
  author={Liu, Ryan Wen and Lu, Yuxu and Gao, Yuan and Guo, Yu and Ren, Wenqi and Zhu, Fenghua and Wang, Fei-Yue},
  journal={IEEE Trans. Intell. Transp. Syst.},
  volume={25},
  number={12},
  pages={19979--19994},
  year={2024},
  publisher={IEEE}
}

@article{liu2021enhanced,
  title={An enhanced CNN-enabled learning method for promoting ship detection in maritime surveillance system},
  author={Liu, Ryan Wen and Yuan, Weiqiao and Chen, Xinqiang and Lu, Yuxu},
  journal={Ocean Eng.},
  volume={235},
  pages={109435},
  year={2021},
  publisher={Elsevier}
}

@inproceedings{habtemariam2012measurement,
  title={Measurement level AIS/radar fusion for maritime surveillance},
  author={Habtemariam, Biruk K and Tharmarasa, R and Meger, Eric and Kirubarajan, T},
  booktitle={Signal and Data Processing of Small Targets 2012},
  volume={8393},
  pages={170--177},
  year={2012},
  organization={SPIE}
}

@article{fukuda2024study,
  title={A study on AIS positional error analysis and transmission frequency requirements of attitude data for future vessel monitoring},
  author={Fukuda, Gen and Tamaru, Hitoi and Kubo, Nobuaki and Shoji, Ruri},
  journal={The Journal of Navigation},
  volume={77},
  number={5-6},
  pages={677--694},
  year={2024},
  publisher={Cambridge University Press}
}

@article{tian2009coordinate,
  title={Coordinate conversion and tracking for very long range radars},
  author={Tian, Xumin and Bar-Shalom, Yaakov},
  journal={IEEE Transactions on Aerospace and Electronic Systems},
  volume={45},
  number={3},
  pages={1073--1088},
  year={2009},
  publisher={IEEE}
}

@inproceedings{singh2020machine,
  title={Machine learning-assisted anomaly detection in maritime navigation using AIS data},
  author={Singh, Sandeep Kumar and Heymann, Frank},
  booktitle={2020 IEEE/ION Position, Location and Navigation Symposium (PLANS)},
  pages={832--838},
  year={2020}
}

@article{kalman1960new,
  title={A new approach to linear filtering and prediction problems},
  author={Kalman, Rudolph Emil},
  journal={Journal of Fluids Engineering},
  volume={82},
  number={1},
  pages={35--45},
  year={1960}
}

@incollection{schmidt1966application,
  title={Application of state-space methods to navigation problems},
  author={Schmidt, Stanley F},
  booktitle={Advances in control systems},
  volume={3},
  pages={293--340},
  year={1966},
  publisher={Elsevier}
}

@article{lu2026cmivtp,
  title={CmIVTP: Cross-modal Interaction-based Vessel Trajectory Prediction for Maritime Intelligence},
  author={Lu, Yuxu and Yang, Dong and Li, Xiaoyu and Bao, Mengwei and Zhao, Congcong},
  journal={arXiv preprint arXiv:2605.26524},
  year={2026}
}



\end{document}